\pdfoutput=1
\PassOptionsToPackage{prologue,dvipsnames}{xcolor}
\documentclass[11pt]{article}

\usepackage[]{acl}

\usepackage{times}
\usepackage{graphicx}
\usepackage{latexsym}
\usepackage{amsmath}
\usepackage{booktabs}
\usepackage{multirow}
\usepackage{enumitem}
\usepackage{xcolor}
\usepackage{color, colortbl}
\usepackage{amssymb}
\usepackage[T1]{fontenc}
\usepackage{algorithm}
\usepackage{algorithmic}
\usepackage[utf8]{inputenc}

\usepackage{microtype}

\usepackage{inconsolata}
\usepackage{adjustbox}
\usepackage{nicematrix}
\usepackage{tabularx}
\usepackage{subcaption}
\usepackage{tikz}
\usepackage{rotating}
\usepackage{makecell}
\usepackage{bbm}
\usepackage{listings}
\usepackage{xspace}
\definecolor{mygray}{gray}{.9}
\definecolor{gray2}{gray}{.78}
\definecolor{gray3}{gray}{.7}
\definecolor{gray4}{gray}{.6}
\definecolor{gray5}{gray}{.5}

\newcommand{\CC}{\cellcolor{mygray}}
\newtheorem{defn}{\hspace{-1mm} Definition}

\newcommand{\rmnum}[1]{\romannumeral #1}
\newcommand{\Rmnum}[1]{\expandafter\@slowromancap\romannumeral #1@}

\newcommand{\cz}[1]{{#1}}

\newcommand{\lyyf}[1]{{#1}}

\newcommand{\as}{Align-Subgraph\xspace}
\newcommand{\ass}{Align-Subgraphs\xspace}
\newcommand{\asea}{\textsc{AsgEa}\xspace}
\newcommand{\asgnn}{\textsc{AsGnn}\xspace}
\newcommand{\ours}{\textsc{AsgEa}-MM\xspace}
\newcommand{\asg}{ASG\xspace}
\newcommand{\asgs}{ASGs\xspace}
\usepackage{tcolorbox}
\newtcbox{\hlprimarytab}{on line, rounded corners, box align=base, colback=c3!10,colframe=white,size=fbox,arc=3pt, before upper=\strut, top=-2pt, bottom=-4pt, left=-2pt, right=-2pt, boxrule=0pt}
\newtcbox{\hlsecondarytab}{on line, box align=base, colback=red!10,colframe=white,size=fbox,arc=3pt, before upper=\strut, top=-2pt, bottom=-4pt, left=-2pt, right=-2pt, boxrule=0pt}
\newcommand{\daugshifted}{\raisebox{0.5\depth}{$\uparrow$}}

\newcommand{\daulg}[1]{{\hlsecondarytab{\daugshifted{#1}}}}

\definecolor{uclablue}{rgb}{0.15, 0.45, 0.68}
\usepackage[fixed]{fontawesome5}
\usepackage{hyperref}

\title{\asea: Exploiting Logic Rules from Align-Subgraphs for \\Entity Alignment}

\author{
  Yangyifei Luo$^{\heartsuit \clubsuit}$, 
  Zhuo Chen$^{\spadesuit}$,
  Lingbing Guo$^{\spadesuit}$,
  Qian Li$^{\heartsuit \clubsuit}$,\\
  \textbf{Wenxuan Zeng$^\heartsuit$,
  Zhixin Cai$^\heartsuit$,
  Jianxin Li$^{\heartsuit \clubsuit}$\thanks{\quad Corresponding Author.}}\\
  $^\heartsuit$School of Computer Science and Engineering, Beihang University  \\
  $^\spadesuit$College of Computer Science and Technology, Zhejiang University \\
  $^\clubsuit$Bejing Advanced Innovation Center for Big Data and Brain Computing, Beihang University \\
 \texttt{ 
    \{luoyyf,liqian,lijx\}@act.buaa.edu.cn,\{zhuo.chen,lbguo\}@zju.edu.cn
  }\\
  \faGithub \textbf{\tt \url{https://github.com/lyyf2002/ASGEA}}
}
\begin{document}
\maketitle
\begin{abstract}
Entity alignment (EA) aims to identify entities across different knowledge graphs that represent the same real-world objects. \cz{Recent embedding-based EA methods have achieved state-of-the-art performance in EA yet faced interpretability challenges as they purely rely on the embedding distance and neglect the \lyyf{logic rules behind a pair of aligned entities}.
In this paper, we propose the Align-Subgraph Entity Alignment (\asea) framework to \lyyf{exploit logic rules from \ass. \asea uses anchor links as bridges to construct \ass and spreads along the paths across KGs, which distinguishes it from the embedding-based methods}. Furthermore, we design an interpretable Path-based Graph Neural Network, \asgnn, to effectively identify and integrate the logic rules across KGs.}  
We also introduce a node-level multi-modal attention mechanism \cz{coupled with multi-modal enriched anchors to augment the \as}. Our experimental results demonstrate the superior performance of \asea~over the existing embedding-based methods in both EA and \cz{Multi-Modal EA (MMEA)} tasks.

\end{abstract}

\section{Introduction}
\begin{figure}[!htbp]
  \centering
  \includegraphics[width=0.99\linewidth]{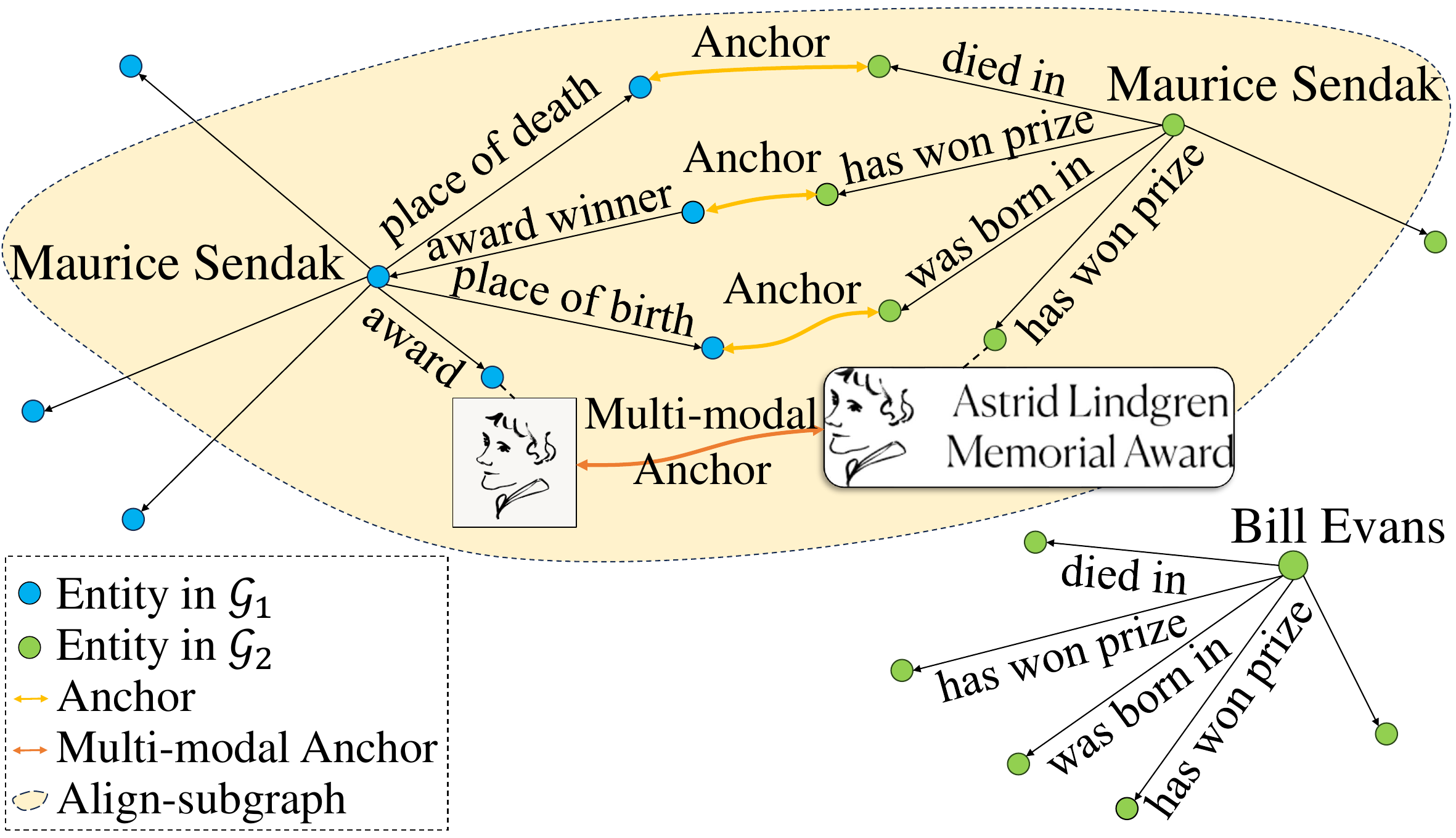}
    \caption{
    \cz{\textbf{\textit{(\rmnum{1})}} 
    The entities ``\textit{Maurice Sendak}'' in $\mathcal{G}_1$ and ``\textit{Bill Evans}'' in $\mathcal{G}_2$ exhibit structurally similar neighbor relationships but they are not candidates for alignment.  
    \textbf{\textit{(\rmnum{2})}} Anchors specifically link ``\textit{Maurice Sendak}'' instances across the KGs without connecting ``\textit{Bill Evans}'' in $\mathcal{G}_2$ to ``\textit{Maurice Sendak}'' in $\mathcal{G}_1$. 
    \textbf{\textit{(\rmnum{3})}} Our \asea~is employed to filter out non-relevant neighbors, retaining only essential alignment information .}
    }
  \label{fig:intro}
  \vspace{-8pt}
\end{figure}
Knowledge Graphs (KGs) provide structured knowledge from real-world facts but suffer from inherent incompleteness. Entity Alignment (EA) addresses this problem by aligning entities across various KGs, facilitating knowledge integration.

Recent advancements in EA have pivoted towards Embedding-Based EA methods \citep{DBLP:conf/ijcai/SunHZQ18, DBLP:conf/acl/CaoLLLLC19}, utilizing vector embeddings of entities and relations to assess similarity. 
Notably, approaches employing  Graph Neural Networks (GNNs)~\citep{DBLP:conf/emnlp/LiuCPLC20, DBLP:conf/kdd/GaoLW0W022} for encoding focus on graph structural representation by neighbor aggregation, positing that entities with analogous neighborhoods are more likely aligned.
However, these methods struggle in scenarios with similar neighborhood structures but different alignment relevance, as shown in  Figure~\ref{fig:intro}. 
This reveals a critical limitation: the uniform representation across entities inadequately distinguishes between relevant and irrelevant neighbors for precise alignment, pointing to the inefficiency of current methods in detailed entity distinction.

Intuitively, considering whether corresponding neighbors are aligned significantly enhances the model's ability to distinguish entities with similar neighborhoods. This is grounded in the fact that if two entities are aligned, their neighbors connected by the same relations are likely aligned as well. By treating alignment seeds as anchor links, such information can be integrated into the model.

Upon further analysis, we find that this essentially establishes a simple logic rule, framed by two identical relations surrounding an alignment relation. Notably, if the spouses of two entities are aligned (the same person), then the entities themselves are also aligned (the same person). However, existing methods overlook the cross-graph logic rules (i.e., alignment rules) hidden behind EA.

To overcome the above challenges, we propose a novel \underline{\textbf{A}}lign-\underline{\textbf{S}}ub\underline{\textbf{g}}raph \underline{\textbf{E}}ntity \underline{\textbf{A}}lignment (\asea) framework. For alignment rule mining, we design an \as~(\asg) extraction algorithm to get specific \asgs~for each entity pair which incorporates the entirety of potential alignment rules between them to filter out non-relevant neighbor information. We further design an \underline{\textbf{A}}lign-\underline{\textbf{S}}ubgraph \underline{\textbf{G}}raph \underline{\textbf{N}}eural \underline{\textbf{N}}etwork (\asgnn) 
which leverages an interpretable attention mechanism to prioritize edges along significant paths and employs a unidirectional Path-based message passing strategy to preserve path-specific information. This ensures the generation of entity-specific representations that focus solely on pivotal alignment details.
To accommodate multi-modal EA (MMEA) tasks, we further innovate a node-level multi-modal attention mechanism. This enhancement is crucial, especially considering the real-world scarcity of anchor links which complicates the acquisition of extensive potential paths for \asgs. By creating new multi-modal anchor links and integrating auxiliary anchor data, we facilitate the extraction of \asgs.
Overall, our contributions can be summarized as:

\begin{figure*}[!htbp]
  \centering
  \includegraphics[width=0.92\linewidth]{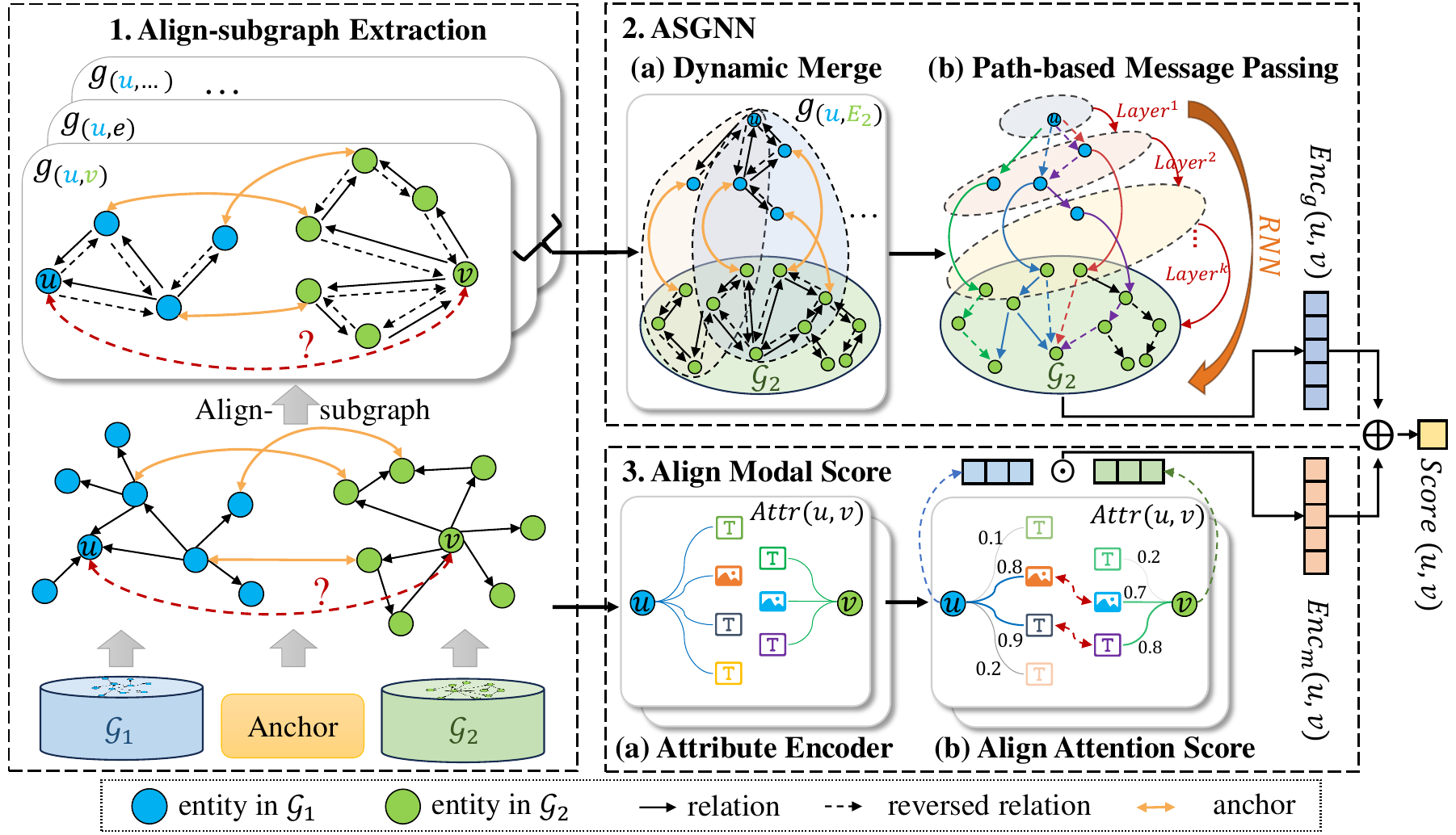}
  \caption{The overall framework of \asea. On the left, the \asg~Extraction is utilized for constructing entity-specific \ass~during training and testing. On the right, the \asgnn~is designed for mining alignment rules and scoring, while the Align modal score is employed for matching aligned attributes to subsequently obtain node-level multi-modal scores.
  }

  \label{fig:model}
  \vspace{-5pt}
\end{figure*}

\begin{itemize}
\item We propose a \asg~extraction algorithm to \cz{ identify} subgraphs \cz{encapsulate} all possible alignment rule paths, \cz{effectively reducing noise from non-relevant neighbors during the alignment process}. Additionally, our \asgnn, an interpretable Path-based GNN for rule mining in \asgs, leverages dynamic programming to improve efficiency in message passing.
\item \cz{Our model is expanded through a node-level unified multi-modal attention mechanism, enhancing its utility for MMEA tasks.}
\item Our \asea~framework achieves \cz{SOTA} performance in both EA and MMEA tasks.
\end{itemize}

\section{Related work}
\subsection{Entity Alignment}

EA integrates KGs by matching entities across graphs, leveraging embedding-based methods to overcome heterogeneity \cite{DBLP:journals/tkde/SunHWWQ23}. These methods fall into two categories: translation-based approaches, employing techniques like TransE \cite{DBLP:conf/nips/BordesUGWY13} to capture structural information within triples \cite{DBLP:conf/ijcai/ChenTYZ17, DBLP:conf/ijcai/ZhangSHCGQ19}, and GNN-based approaches, using networks such as GCN \cite{DBLP:conf/iclr/KipfW17} to aggregate neighborhood features for richer entity representations \cite{DBLP:conf/acl/CaoLLLLC19, DBLP:conf/emnlp/ShiX19}. Enhancements include shared parameter embeddings for seed alignment consistency \cite{DBLP:conf/ijcai/ZhuXLS17} and iterative learning~\cite{DBLP:conf/ijcai/SunHZQ18}. PEEA's positional encoding introduces global relational perspectives, but it remains reliant on embedding techniques \cite{DBLP:conf/wsdm/TangS00WTY23}.

\subsection{Multi-modal Entity Alignment}
In the emerging field of MMEA, researchers address the complex nature of real-world KGs by integrating multiple modalities \cite{DBLP:journals/tkde/ZhuLWJSWXY24,chen2024knowledge}. 
\citet{DBLP:conf/aaai/0001CRC21} introduce attention mechanisms to weigh modalities differently, while~\citet{DBLP:conf/kdd/ChenL00WYC22} and~\citet{DBLP:conf/coling/LinZWSW022} focus on integrating visual cues and enhancing intra-modal learning, respectively.~\citet{DBLP:conf/mm/ChenCZGFHZGPSC23} adopts a transformer-based multi-modal approach, and~\citet{DBLP:conf/www/LiGLJWSL23} targets the contextual gap by ensuring attribute consistency across modalities.

Recent rule-based EA methods \cite{DBLP:journals/pvldb/LeoneHAGW22, DBLP:conf/icassp/LiCLXCZ23} primarily build upon the PARIS \cite{DBLP:journals/pvldb/SuchanekAS11}, a probabilistic approach that leverages relationship functionality. However, the lack of public access to code or the use of differing metrics hampers direct comparisons with \asea.
To the best of our knowledge, we are the first to employ logic rules for EA.
\subsection{Rule-based Reasoning}
Knowledge graph reasoning leverages entity and relationship analysis to infer new knowledge\cite{DBLP:journals/corr/abs-2212-05767}. Various methods have been developed for reasoning through logic rule mining. AIME~\cite{DBLP:conf/www/GalarragaTHS13} initiates the exploration with inductive logic programming for rule discovery in KGs. Neural LP~\cite{DBLP:conf/nips/YangYC17} and DRUM~\cite{DBLP:conf/nips/SadeghianADW19} later apply dynamic programming to create linear combinations of logic rules. AnyBURL~\cite{DBLP:conf/ijcai/MeilickeCRS19} introduces a bottom-up approach for efficient rule learning in large graphs. RED-GNN~\cite{DBLP:conf/www/ZhangY22} adopts a Bellman-Ford-like method for learning path representations. 
Rlogic~\cite{DBLP:conf/kdd/ChengL0S22}, employs a deductive reasoning-based scoring model for quality evaluation of rules derived from sampled paths. However, these methods mainly learn logic rules within a single KG, not addressing EA across multiple KGs.

\section{Methodology}
\subsection{Preliminaries}
Following \citet{DBLP:conf/semweb/ChenGFZCPLCZ23,DBLP:conf/mm/ChenCZGFHZGPSC23},
we define a MMKG as $\mathcal{G} = \{\mathcal{E}, \mathcal{R}, \mathcal{A}, \mathcal{V}, \mathcal{T}\}$, where $\mathcal{E}$, $\mathcal{R}$, $\mathcal{A}$, and $\mathcal{V}$ denote the sets of entities, relations, attributes, and attribute values, respectively. The set $\mathcal{T}$ is divided into $\mathcal{T}_\mathcal{R}$ and $\mathcal{T}_\mathcal{A}$, where $\mathcal{T}_\mathcal{R} \subseteq \mathcal{E} \times \mathcal{R} \times \mathcal{E}$ is the set of relation triples, and $\mathcal{T}_\mathcal{A} \subseteq \mathcal{E} \times \mathcal{A} \times \mathcal{V}$ is the set of attribute triples. For traditional KG, the sets $\mathcal{A}$, $\mathcal{V}$, and $\mathcal{T}_\mathcal{A}$ are empty.
Given two MMKGs $\mathcal{G}_1$ and $\mathcal{G}_2$, the goal of MMEA is to identify pairs of entities $(e^1_i, e^2_i)$, where $e^1_i\in\mathcal{E}_1, $$e^2_i\in\mathcal{E}_2$ that represent the same real-world entity $e_i$. 
A set of aligned entity pairs is provided, which is partitioned into a training subset (i.e., seed alignments $\mathcal{S}$) and a testing subset $\mathcal{S}_{te}$, according to a specified seed alignment ratio ($R_{sa}$).
\begin{defn}{\textbf{Anchor Link.}}
For $e_u\in\mathcal{E}_1 $ and $e_v\in\mathcal{E}_2$, an anchor link between them is denoted as $A(e_u,e_v)$, which means $e_u$ is aligned with $e_v$. 
\end{defn}
\subsection{\as~Extraction}

\subsubsection{\textbf{Alignment Rules}}\label{sec:rules} 

\begin{figure*}[!htbp]\label{fig:rule}
    \centering
    \begin{subfigure}[b]{0.32\textwidth}
        \includegraphics[width=\textwidth]{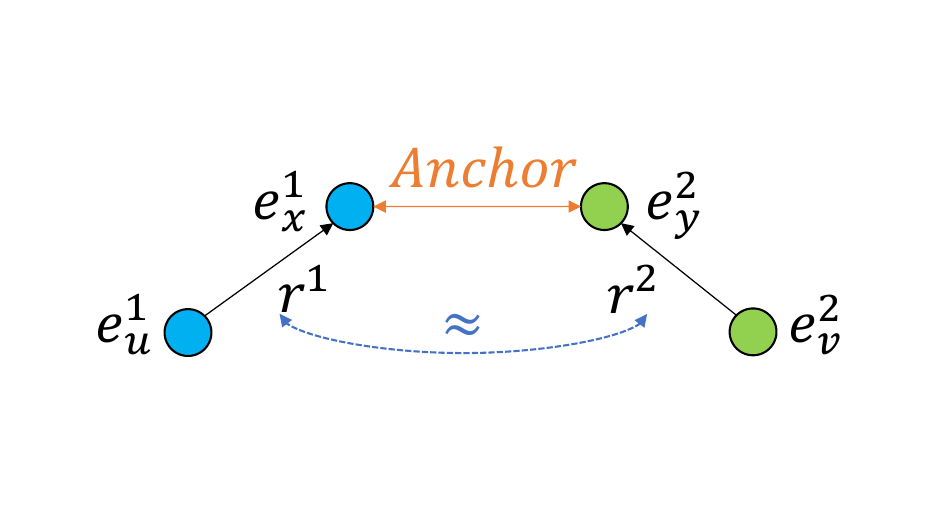}
        \caption{One-hop rule}
        \label{fig:rule1}
    \end{subfigure}
    \hfill
    \begin{subfigure}[b]{0.32\textwidth}
        \includegraphics[width=\textwidth]{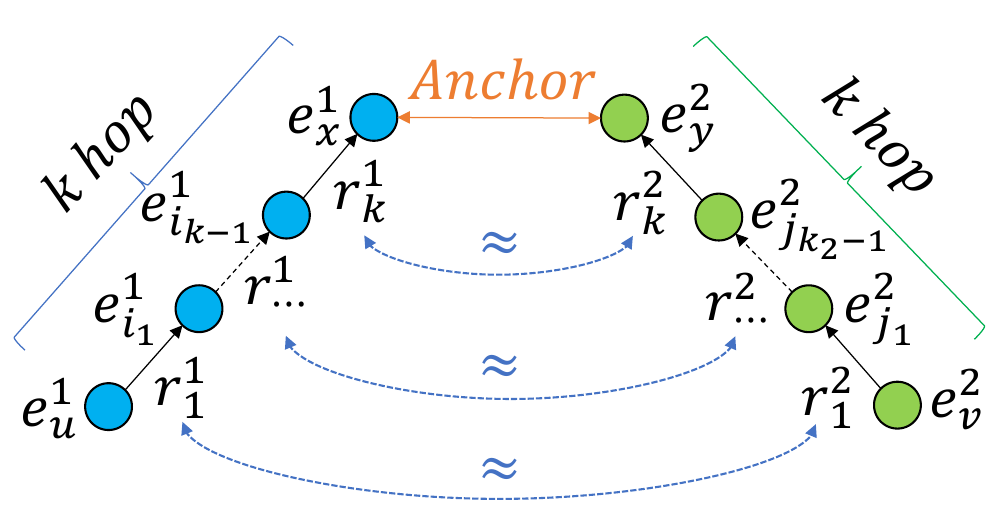}
        \caption{Symmetric k-hop rule}
        \label{fig:rule2}
    \end{subfigure}
    \hfill
    \begin{subfigure}[b]{0.32\textwidth}
        \includegraphics[width=\textwidth]{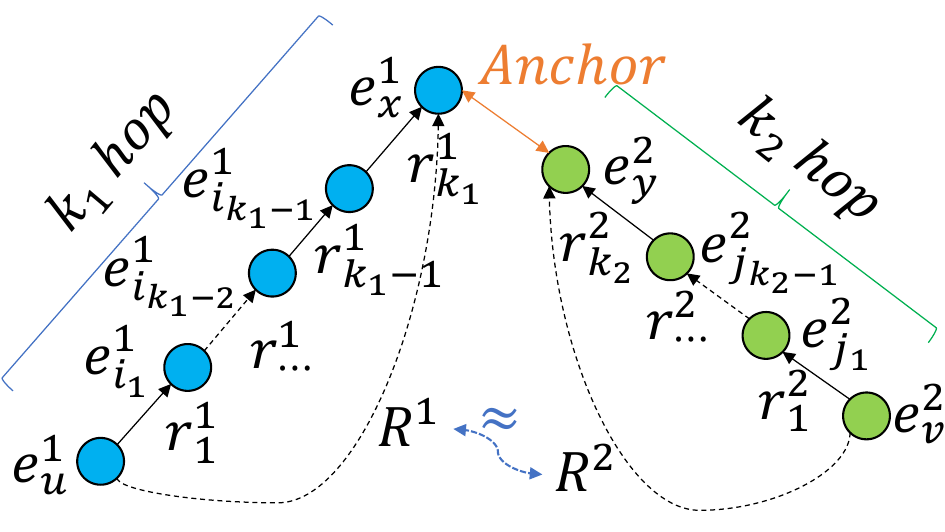}
        \caption{Asymmetric k-hop rule}
        \label{fig:rule3}
    \end{subfigure}
    \caption{Illustration of different alignment rules.}
    \vspace{-5pt}
\end{figure*}

Alignment rules are logical constructs that identify aligned entities across KGs using structural and semantic information. We introduce three rules:

\textbf{One-hop Rule}: As shown in Figure \ref{fig:rule1}, for entities \(e_u^1\) and \(e_v^2\) with neighbors \(e_x^1\) and \(e_y^2\) in KGs \(\mathcal{G}_1\) and \(\mathcal{G}_2\) respectively, if \(e_x^1\) and \(e_y^2\) are aligned (\(A(e_x^1, e_y^2)\)) and the relations \(r^1(e_u^1, e_x^1)\) and \(r^2(e_v^2, e_y^2)\) are identical, then \(e_u^1\) and \(e_v^2\) are aligned. This is summarized as \(A(e_u^1,e_v^2) \leftarrow r^1(e_u^1,e_x^1) \land r^2(e_v^2,e_y^2) \land r^1=r^2\land A(e_x^1,e_y^2)\).

\textbf{Symmetric k-hop Rule}: As shown in Figure \ref{fig:rule2}, \(e_u^1\) and \(e_v^2\) are connected to their respective \(k\)-hop neighbors \(e_x^1\) and \(e_y^2\) through identical sequences of relations. \(e_u^1\) and \(e_v^2\) are aligned if \(e_x^1\) and \(e_y^2\) are aligned, which is expressed as \(A(e_u^1, e_v^2)\leftarrow r_1^1(e_u^1, e_{i_1}^1) \land \ldots \land r_k^1(e_{i_{k-1}}^1, e_x^1) \land r_1^2(e_v^2, e_{j_1}^2) \land \ldots \land r_k^2(e_{j_{k-1}}^2, e_y^2) \land r_1^1 = r_1^2 \land \ldots \land r_k^1 = r_k^2 \land A(e_x^1, e_y^2)\).

\textbf{Asymmetric k-hop Rule}: In Figure \ref{fig:rule3}, \(e_u^1\) and \(e_v^2\) connected to \(e_x^1\) and \(e_y^2\) through different numbers of hops. If the composite relations \(R^1\) and \(R^2\) between \(e_u^1\) and \(e_x^1\), and \(e_v^2\) and \(e_y^2\) are semantically equivalent and \(e_x^1\), \(e_y^2\) are aligned, then \(e_u^1\) and \(e_v^2\) are aligned, which is expressed as \(A(e_u^1, e_v^2)\leftarrow R^1(e_u^1, e_x^1) \land R^2(e_v^2, e_y^2) \land R^1 = R^2 \land A(e_x^1, e_y^2)\).

\subsubsection{\textbf{Alignment Path}} 
\cz{Expanding on the Alignment Rule analysis, we integrate anchor link relations
($r_{anchor}$) to connect the two KGs. We categorize anchor links into:
\textbf{\textit{(\rmnum{1})} Anchor}: Utilizes established alignment seeds as anchors. Training strategies are detailed in \S~\ref{train strategy}.
\textbf{\textit{(\rmnum{2})} Multi-modal Anchor}: 
Forms connections between nodes with similar multi-modal attributes.
}

We reverse relations within KGs (remarked as $r'$) for path-finding. The alignment rule in \S~\ref{sec:rules} is rewritten as \( R^1(e_u^1, e_x^1) \land r_{anchor}(e_x^1, e_y^2) \land {R^2}'(e_y^2, e_v^2) \rightarrow A(e_u^1, e_v^2) \). Based on this rule, we define the Alignment Path below:

\begin{defn}{\textbf{Alignment Path.}}
An alignment path \(P(e_u^1, e_v^2)\) with length $L$ is a sequence of relations $e_u^1 \xrightarrow{r_1^1} \ldots \xrightarrow{r_{k_1}^1} e_x^1 \xrightarrow{r_{anchor}} e_y^2 \xrightarrow{{r_{k_2}^2}'} \ldots \xrightarrow{{r_{1}^2}'} e_v^2$ from $e_u^1$ to $e_v^2$ that follows alignment rules.
\end{defn}

\subsubsection{\textbf{\as}} 

Considering that a solitary alignment path might not suffice for definitive entity alignment, it's crucial to aggregate all potential alignment paths between entities \( e_u \) and \( e_v \). Thus, we define an \as~(\asg) for a pair \( (e_u, e_v) \).
\begin{defn}{\textbf{\as}.}
For entity pair \( (e_u, e_v) \), \as~\( g(e_u, e_v) \) is the subgraph comprised of all alignment paths between \( e_u \) and \( e_v \), formally denoted as \( g(e_u, e_v) = \bigcup P(e_u, e_v) \).
\end{defn}

\begin{defn}{\textbf{$K$-hop \as}.}
$K$-hop \as~is an \as~formed by integrating all alignment paths with length \( \le K \), formally denoted as \( g^K(e_u, e_v) = \bigcup\nolimits_{i \leq K} P^i(e_u, e_v) \).
\end{defn}

We designed an \asg~extraction algorithm to obtain the $k$-hop \asg~for the entity pair $(e_u,e_v)$, see Appendix for more details.

\subsection{ASGNN}
\subsubsection{\textbf{Dynamic Merge}} 
\begin{figure}[!htbp]
  \centering
  \includegraphics[width=0.9\linewidth]{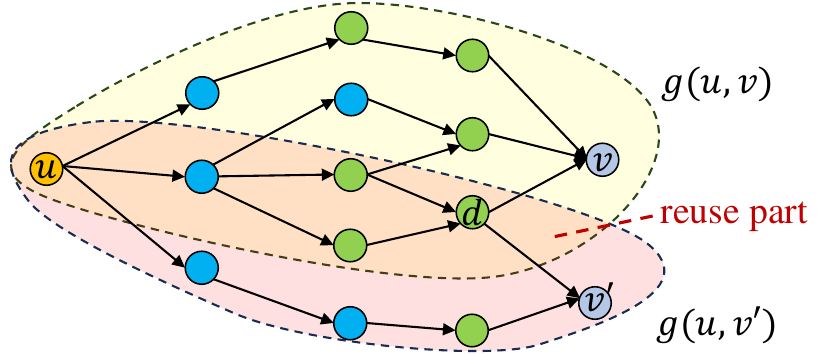}
  \caption{An example for dynamic merge. The yellow portion denotes \asg~\( g(e_u, e_v) \) while the red portion denotes \asg~\( g(e_u, e_v') \). The overlapping area signifies the parts that can be reused through dynamic programming.
  }
  \label{fig:merge}
  \vspace{-5pt}
\end{figure}

To address the computational challenge of getting alignment scores between an entity \( e_u \) from $\mathcal{G}_1$ and all entities in $\mathcal{G}_2$, we employ a dynamic merge strategy, inspired by \citet{DBLP:conf/www/ZhangY22}.

Our method facilitates unidirectional message passing from \(e_u\) towards entities in \(\mathcal{G}_2\), including specific instances like \(e_v\) and \(e_{v'}\) as shown in Figure \ref{fig:merge}. When \(e_v\) and \(e_{v'}\) share a common ancestor \(d\), we utilize dynamic programming to share \(d\)'s representation, significantly reducing computational efforts by merging convergent alignment paths.

To evaluate \( e_u \)'s alignment with all entities in $\mathcal{G}_2$, we construct a merged \asg~\(g(e_u, \mathcal{E}_2)\) encapsulating all paths from \(e_u\) to \( \mathcal{E}_2 \), formalized as:
\begin{equation}
    g(e_u,\mathcal{E}_2)=\bigcup\nolimits_{e_v\in \mathcal{E}_2}g(e_u,e_v)\label{eq:gue2}.
\end{equation}

\subsubsection{\textbf{Path-based Message Passing}} 

We introduce a Path-based Message Passing technique within an \as~Graph Neural Network (\asgnn) to facilitate directed information flow from entity \(e_u\) to \( \mathcal{E}_2 \). In the merged \asg~\(g^L(e_u, \mathcal{E}_2)\), we ensure each node, apart from \(e_u\), is involved in at least one alignment path, thus incorporating incoming edge information effectively. The computation of the path representation from \(e_u\) to any entity \(e_i\) is defined as follows:
\begin{equation}
    \smash{h_{p(e_u,e_i)}^l=\sum\nolimits_{e_j \in N(e_i)} \alpha^{e_u,l}_{e_j,e_i} (h^{l-1}_{e_u,e_j} + r^l_{e_j,e_i}), }
\end{equation}
where \(\smash{ e_j \in N(e_i)} \) denotes the neighbors with edges pointing to entity \( e_i \), and \( \smash{r^l_{e_j,e_i}} \) represents the relation representation between \( e_j\) and \(e_i\). \( \smash{\alpha^{e_u,l}_{e_j,e_i}} \) is determined via an interpretable attention mechanism, using an MLP to prioritize the most significant alignment paths:
\begin{equation}\label{eq:alpha}
    \alpha^{e_u,l}_{e_j,e_i} = \text{MLP}_{\alpha}^l(h^{l-1}_{e_u,e_j} \, || \, r^l_{e_j,e_i} \, || \, f_{g}(e_j,e_i)),
\end{equation}
where \(f_{g}(e_j,e_i)\) incorporates graph knowledge, differentiating edges within or across KGs to accommodate their heterogeneity, calculated as:
\begin{equation}
    f_{g}(e_j,e_i) = \text{MLP}_{kg}(\text{EMB}(e_j) \, || \, \text{EMB}(e_i)), 
\end{equation}
where EMB is a learnable embedding function.We employ an RNN for information updates
to preserve crucial path information:
\begin{equation}
   \smash{ h^{l}_{e_u,e_i} = \text{RNN}(h^{l-1}_{e_u,e_i}, W^l h_{p(e_u,e_i)}^l) }.
\end{equation}

After \( L \) layers of updates on \( g^L(e_u, \mathcal{E}_2) \), we obtain the final representations \( h^L_{e_u,e_v} \) for all \( e_v \in \mathcal{E}_2 \) which encapsulate critical alignment path information for pair \( (e_u,e_v) \). \cz{We employ a MLP for scoring}, defined as \( \text{score}_g(e_u,e_v) = \text{MLP}_g(h^L_{e_u,e_v}) \).

\subsection{Align Modal Score}
\subsubsection{\textbf{Attribute Encoder}} 

To integrate text and image attributes from MMKGs into a unified representation space, we utilize modality-specific fully connected layers, \(W_\mathcal{M} \in \mathbb{R}^{d_\mathcal{M} \times d}\), to project features into a unified space via \(h_{\mathcal{A}}(v) = W_\mathcal{M} h_{\mathcal{A}}^{0}(v)\). Initial embeddings \(h_{\mathcal{A}}^{0}(v)\) are derived from BERT for text and a pre-trained model for images.

\subsubsection{\textbf{Align Attention Score}} 
To mitigate the noise from varying attribute types and quantities in EA, we introduce an align attention mechanism, which prioritizes attributes shared between entities \(e_u\) and \(e_v\), diminishing the influence of mismatched attributes. We calculate pair-aware multi-modal representations \(h^{e_u,e_v}_u\) as:
\begin{equation}
    h^{e_u,e_v}_u = \sum_{e_u, a_i, v_i \in \mathcal{T}_\mathcal{A}(e_u)} \alpha^{e_u,e_v}_{a_i} h_{\mathcal{A}}(v_i).
\end{equation}

Attention scores \( \alpha^{e_u,e_v}_{a_i} \) amplify the weights of more relevant attributes, calculated using a softmax function over the similarity between attribute types:
\begin{equation}
\alpha^{e_u,e_v}_{a_i} = \sigma\left( \sum\nolimits_{e_v, a_j, v_j \in \mathcal{T}_\mathcal{A}(e_v)} \text{sim}(a_i, a_j) \right).
\end{equation}

Finally, the multi-modal attribute alignment score, which quantifies the similarity between \(e_u\) and \(e_v\), is determined by processing their combined multi-modal representations through an MLP: \(\text{score}_m(e_u,e_v) = \text{MLP}_m(h^{e_u,e_v}_u \odot h^{e_v,e_u}_v)\).
\subsection{Training Objective}
\subsubsection{Training Strategy}\label{train strategy}
During the test phase, we utilize all alignment seeds as anchor links to extract \ass~for evaluation. To emulate this process in training, we randomly split the alignment seeds \(S\) into two subsets: \(S_{anchor}\) (75\%) and \(S_{train}\) (25\%). \(S_{anchor}\) acts as known anchor links to extract \ass~for predicting the remaining \(S_{train}\) subset.
\subsubsection{Loss Function}

We present two variants: \asea-Stru, focusing on graph structure, and \asea-MM, incorporating multimodal data. \asea-Stru only uses \(\text{score}_g(e_u,e_v)\) from \asgnn for alignment scores, while \asea-MM adds a multimodal score, yielding \(s(e_u,e_v) = \text{score}_g(e_u,e_v) + \text{score}_m(e_u,e_v)\).

We treat \(e_v\), the entity aligned with \(e_u\), as a positive example against all other entities as negatives. For each pair of \(S_{train}\), we utilize the following loss for optimization \cite{DBLP:conf/icml/LacroixUO18}:
\begin{equation}
   \smash{ l(e_u) = \log\left(\sum\nolimits_{e\in \hat{\mathcal{G}}}e^{s(e_u,e)}\right)-s(e_u,e_v) },
\end{equation}
\begin{equation}
    \smash{\mathcal{L} = \sum\nolimits_{(e_u,e_v) \in S_{train}}(l(e_u)+l(e_v))}\label{eq:loss},
\end{equation}
where \(\hat{\mathcal{G}}\) represents the KG opposite to the one containing \(e_u\). We use stochastic gradient descent~\cite{DBLP:journals/corr/KingmaB14} to minimize \eqref{eq:loss}.

\section{Experiment}
\subsection{Dataset}
\label{sec:dataset}

\textbf{\textit{(\rmnum{1})}} MMKG~\cite{DBLP:conf/esws/LiuLGNOR19} include two subsets: FB15K-DB15K (FBDB15K) and FB15K-YAGO15K (FBYG15K). These datasets are split into three subsets based on the ratio of seed alignments: $R_{sa} \in \{0.2, 0.5, 0.8\}$.

\textbf{\textit{(\rmnum{2})}} DBP15K~\cite{DBLP:conf/semweb/SunHL17} are derived from the multilingual versions of DBpedia and include three subsets: DBP15K$_{\text{ZH-EN}}$, DBP15K$_{\text{JA-EN}}$, and DBP15K$_{\text{FR-EN}}$. We incorporate the multi-modal variant of these datasets \cite{DBLP:conf/aaai/0001CRC21}, which enhances the entities with corresponding images.

\textbf{\textit{(\rmnum{3})}} Multi-OpenEA~\cite{DBLP:conf/icassp/LiCLXCZ23} are the multi-modal variants of the OpenEA benchmarks \cite{DBLP:journals/pvldb/SunZHWCAL20} which enhance entity images through Google search. We use the datasets \{EN-FR-15K, EN-DE-15K, D-W-15K-V1, D-W-15K-V2\}, following \citet{DBLP:conf/semweb/ChenGFZCPLCZ23} to select only the rank one image for each entity.

\subsection{Implementation Details}
We use pre-trained BERT for text attribute initialization and the dimension of text $d_{t}$ is 768. Following \cite{DBLP:conf/ksem/ChenLWXWC20,DBLP:conf/coling/LinZWSW022,DBLP:conf/semweb/ChenGFZCPLCZ23}, the vision encoders are set to ResNet-152 \cite{DBLP:conf/cvpr/HeZRS16} on DBP15K with the vision dimension $d_v =2048$, and set to VGG-16 \cite{DBLP:journals/corr/SimonyanZ14a} on MMKG with $d_v=4096$, and set to CLIP \cite{DBLP:conf/icml/RadfordKHRGASAM21} on Multi-OpenEA with $d_v=512$.

\subsection{Main Results}
\cz{Across all datasets, surface information is excluded for fair comparison,  following \citet{DBLP:conf/semweb/ChenGFZCPLCZ23}.}
\subsubsection{MMKG Dataset}
\begin{figure}[!htbp]
  \centering
  \includegraphics[width=0.99\linewidth]{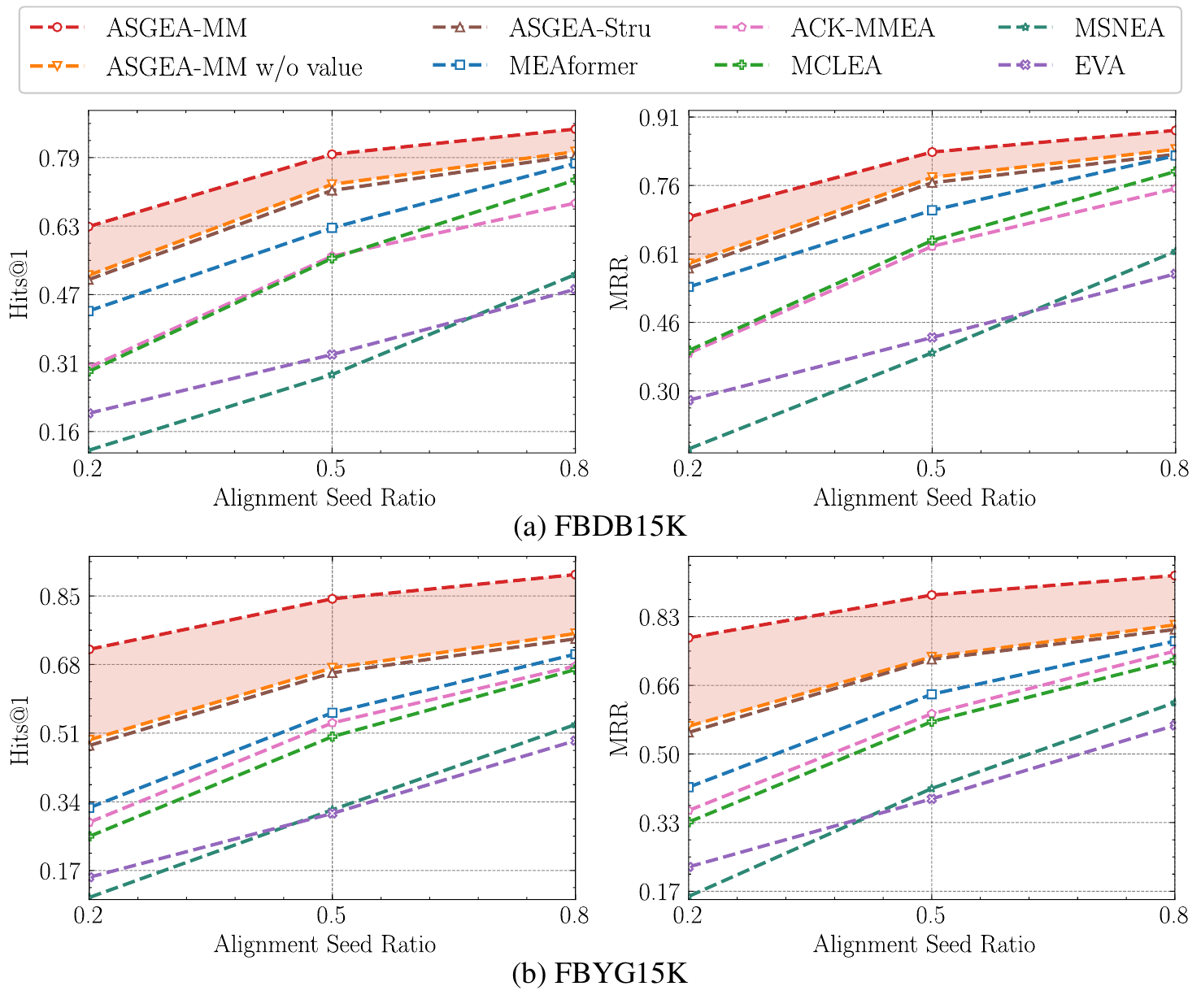}
    \caption{Hits@1 and MRR performance on FBDB15K and FBYG15K with different $R_{sa}$. \textbf{``-MM''} indicates using multi-modal information(abbreviated as MM), while \textbf{``-Stru''} indicates reliance on graph structure alone (i.e., triples). ``w/o value'' indicates without attribute values. For complete results and baseline details, see Appendix.}
  \label{fig:fbdb}

\end{figure}
Results on FBDB15K and FBYG15K datasets are detailed in Figure \ref{fig:fbdb}, providing insights as follows:
\textbf{\textit{(\rmnum{1})}}~\ours~exceeds baselines in all metrics and data proportions, achieving MRRs up to 0.881 on FBDB15K and 0.926 on FBYG15K, demonstrating significant and consistent advantages.
\textbf{\textit{(\rmnum{2})}}~Delving deeper into the results, we observe that in the H@1 metric, \ours~exhibits substantial improvements of \daulg{19.4\%}, \daulg{16.9\%}, and \daulg{7.9\%} for the 20\%, 50\%, and 80\% training data splits on FBDB15K (averaging \textbf{\daulg{14.7\%}}) and \daulg{39.2\%}, \daulg{28.2\%}, and \daulg{19.7\%} on the FBYG15K (averaging \textbf{\daulg{29.0\%}}). In terms of the MRR metric, the model shows a steady enhancement with increases of \daulg{15.5\%}, \daulg{12.9\%}, and \daulg{5.6\%} for the 20\%, 50\%, and 80\% training data proportions on FBDB15K (averaging \textbf{\daulg{11.3\%}}), and \daulg{36.0\%}, \daulg{23.9\%}, and \daulg{15.8\%} on FBYG15K (averaging \textbf{\daulg{25.2\%}}).
\textbf{\textit{(\rmnum{3})}}~\asea-Stru, utilizing solely graph structure data and \textbf{excluding any auxiliary information like images or texts}, outperforms all existing traditional EA and MMEA baselines. This performance indicates our model's strong capability in mining alignment rules. 
In terms of architecture, \asgnn~functions similarly to a subgraph GNN, which is shown to possess greater expressive power than traditional GNN models~\cite{DBLP:conf/nips/FrascaBBM22}.
\textbf{\textit{(\rmnum{4})}}~Previous studies have concentrated on one-hot encoding of attribute types \cite{DBLP:conf/mm/ChenCZGFHZGPSC23}, overlooking the significance of attribute values. To ensure a fair comparison with them, we present a model variant ``w/o value'' that solely relies on attribute types. Our findings demonstrate the robust performance of \asea and underscore the importance of incorporating attribute values.
\textbf{\textit{(\rmnum{5})}}~Considering the absence of models that apply rule-based reasoning techniques to EA tasks, we adapt several reasoning methods Neural LP~\cite{DBLP:conf/nips/YangYC17}, AnyBURL~\cite{DBLP:conf/ijcai/MeilickeCRS19} and RED-GNN~\cite{DBLP:conf/www/ZhangY22} in our framework to tackle EA challenges. The adaptations for the alignment tasks are denoted by the suffix ``$\ast$''. Notably, they achieve commendable results, underscoring the efficacy of \asea.

\subsubsection{DBP15K Dataset}
Results on three datasets of DBP15K with $R_{sa} = 0.3$ following \cite{DBLP:conf/aaai/0001CRC21} are detailed in Table \ref{tab:dbp15k}, \ours~outperforms all other baselines on three datasets in both EA and MMEA settings. Notably, \ours~achieves up to \daulg{9.3\%} improvement in H@1 and \daulg{6.1\%} in MRR.
Different from MMKG Dataset, these models generally perform better than those without multi-modal information, indicating the importance of multi-modal information in MMEA and the potential low quality of images in the MMKG dataset. 
\begin{table}[!htbp]
    \centering
    \tabcolsep=0.3cm
    \renewcommand\arraystretch{1.0}
    \resizebox{0.99\linewidth}{!}{
    \begin{tabular}{@{}l|l|cc|cc|cc}
        \toprule
        & \multirow{2}*{\makebox[2cm][c]{Models}} & \multicolumn{2}{c|}{DBP15K$_{\text{ZH-EN}}$} & \multicolumn{2}{c|}{DBP15K$_{\text{JA-EN}}$} & \multicolumn{2}{c}{DBP15K$_{\text{FR-EN}}$} \\
        & & {\scriptsize H@1} & {\scriptsize MRR} & {\scriptsize H@1} & {\scriptsize MRR} & {\scriptsize H@1} & {\scriptsize MRR} \\
        \midrule
        \parbox[t]{2mm}{\multirow{5}{*}{\rotatebox[origin=c]{90}{EA}}}
        &AlignEA 
        & 
        .472 & .581 & .448 & .563 & .481 & .599 \\
        &KECG 
        &
        .478 & .598 & .490 & .610 & .486 & .610 \\
        &MUGNN 
        &
        .494 & .611 & .501 & .621 & .495 & .621 \\
        &AliNet 
        &
        \underline{.539} & \underline{.628} & \underline{.549} & \underline{.645} & \underline{.552} & \underline{.657} \\
        &\CC\textbf{\asea-Stru}  &
        \CC\textbf{.560} & \CC\textbf{.660} & \CC\textbf{.595} & \CC\textbf{.690} & \CC\textbf{.653} & \CC\textbf{.745} \\
        \midrule
        \parbox[t]{2mm}{\multirow{6}{*}{\rotatebox[origin=c]{90}{MMEA}}}
        &MSNEA 
        & .609 & .685 & .541 & .620 & .557 & .643 \\
        &EVA 
        &
        {.683} & {.762} & {.669} & {.752} & {.686} & {.771} \\
        &MCLEA 
        &
        .726 & .796 & .719 & .789 & .719 & .792 \\
        &MEAformer 
        &
        .771 & .835 & .764 & .834 & .770 & .841 \\
        &UMAEA 
        & \underline{.800} & \underline{.860} & \underline{.801} & \underline{.862} & \underline{.818} & \underline{.877} \\
        &\CC\textbf{\ours}  &
        \CC\textbf{.815} & \CC\textbf{.862} & \CC\textbf{.849} & \CC\textbf{.889} & \CC\textbf{.911} & \CC\textbf{.938} \\
        
        \bottomrule
    \end{tabular}
    }
    \caption{\cz{Results on DPB15K datasets. For complete results and baseline details, see the Appendix.} 
    }
    \label{tab:dbp15k}
\end{table}

\subsubsection{Multi-OpenEA Dataset}
Results on Multi-OpenEA with $R_{sa} = 0.2$ are detailed in Table \ref{tab:moea}.
\ours~surpasses all baseline models in both EN-FR-15K and EN-DE-15K dataset. 
Specifically, \ours~demonstrates an improvement of \daulg{6.3\%} in H@1 and \daulg{4.1\%} in MRR for the EN-FR-15K dataset.

\begin{table}[!htbp]
    \centering
    \tabcolsep=0.3cm
    \renewcommand\arraystretch{1.0}
    \resizebox{0.98\linewidth}{!}{
    \begin{tabular}{@{}l|ccc|ccc}
        \toprule
        \multirow{2}*{\makebox[2cm][c]{Models}} & \multicolumn{3}{c|}{OpenEA$_{\text{EN-FR}}$} & \multicolumn{3}{c}{OpenEA$_{\text{EN-DE}}$} \\
        & {\scriptsize H@1} & {\scriptsize H@10} & {\scriptsize MRR} & {\scriptsize H@1} & {\scriptsize H@10} & {\scriptsize MRR} \\
        \midrule
        MSNEA {\footnotesize {\cite{DBLP:conf/kdd/ChenL00WYC22}}} 
        & .692 & .813 & .734 & .753 & .895 & .804 \\
        EVA {\footnotesize \cite{DBLP:conf/aaai/0001CRC21}} 
        & {.785} & {.932} & {.836} & {.922} & {.983} & {.945}  \\
        MCLEA {\footnotesize {\cite{DBLP:conf/coling/LinZWSW022}}}  
        & .819 & .943 & .864 & .939 & .988 & .957  \\
        UMAEA {\footnotesize {\cite{DBLP:conf/semweb/ChenGFZCPLCZ23}}}  
        & .848 & .966 & .891 & .956 & .994 & .971  \\
        \CC\textbf{\ours}  
        & \CC\textbf{.911} & \CC\textbf{.968} & \CC\textbf{.932} & \CC\textbf{.980} & \CC\textbf{.996} & \CC\textbf{.986} \\

        \bottomrule
    \end{tabular}
    }
    \caption{Results on Multi-OpenEA datasets. For complete results and baseline details, see the Appendix.  
    }
    \label{tab:moea}
\end{table}

\begin{table}[ht]
    \centering
    \renewcommand\arraystretch{1.0}
    \resizebox{0.9\linewidth}{!}{
    \begin{tabular}{@{}l|l|ccc|ccc}
        \toprule
        & \multirow{2}*{\makebox[1.8cm][c]{Models}} & \multicolumn{3}{c|}{FBDB15K} & \multicolumn{3}{c}{FBYG15K} \\
        & & {\scriptsize H@1} & {\scriptsize H@10} & {\scriptsize MRR} & {\scriptsize H@1} & {\scriptsize H@10} & {\scriptsize MRR} \\
        \midrule
        \parbox[t]{1.2mm}{\multirow{4}{*}{\rotatebox[origin=c]{90}{20\%}}} 
        & \CC\textbf{\ours} & \CC\textbf{.628} & \CC\textbf{.799} & \CC\textbf{.689} & \CC\textbf{.717} & \CC\textbf{.848} & \CC\textbf{.776} \\
        & ~~~~w/o AMS & .531 & .734 & .600 & .500 & .705 & .572 \\
        & \asea-Stru & .509 & .712 & .579 & .486 & .688 & .555 \\
        & ~~~~symmetric & .499 & .702 & .569 & .473 & .674 & .542 \\

        \midrule
        \parbox[t]{1.2mm}{\multirow{4}{*}{\rotatebox[origin=c]{90}{50\%}}} 

        & \CC\textbf{\ours} & \CC\textbf{.794} & \CC\textbf{.899} & \CC\textbf{.833} & \CC\textbf{.842} & \CC\textbf{.920} & \CC\textbf{.879} \\
        & ~~~~w/o AMS & .738 & .876 & .788 & .690 & .852 & .749 \\
        & \asea-Stru & .712 & .855 & .764 & .671 & .836 & .730 \\
        & ~~~~symmetric & .701 & .848 & .753 & .659 & .823 & .717 \\

        \midrule
        \parbox[t]{1.2mm}{\multirow{4}{*}{\rotatebox[origin=c]{90}{80\%}}} 
        & \CC\textbf{\ours} & \CC\textbf{.852} & \CC\textbf{.927} & \CC\textbf{.881} & \CC\textbf{.902} & \CC\textbf{.965} & \CC\textbf{.926} \\
        & ~~~~w/o AMS & .818 & .916 & .854 & .796 & .916 & .839 \\
        & \asea-Stru & .803 & .904 & .841 & .758 & .900 & .808 \\
        & ~~~~symmetric & .788 & .891 & .825 & .747 & .885 & .798 \\

        \bottomrule
    \end{tabular}
    }
    \caption{Variant experiment results on MMKG datasets. \textbf{``\asea-MM''} indicates our complete model that uses multi-modal information(abbreviated as MM), while \textbf{``\asea-Stru''} indicates reliance on graph structure alone (i.e., triples), \textbf{``w/o AMS''} denotes the exclusion of the AMS module from the complete model, and \textbf{``symmetric''} indicates employing only symmetric alignment rule based on the variant ``\asea-Stru''.
    }
    \label{tab:FBDBYG-Variant}
    \vspace{-4pt}
\end{table}
\begin{table*}[!htbp]
    \centering
    \renewcommand\arraystretch{1.0}
    \resizebox{0.95\linewidth}{!}{
    \begin{tabular}{@{}l|l|ccc|ccc}
        \midrule
        \parbox[t]{1.2mm}{\multirow{3}{*}{\rotatebox[origin=c]{90}{hop=3}}} 
        & \texttt{A(X,Y)} \(\leftarrow\) \texttt{directed\_by(A,X)} \(\wedge\) \texttt{A(A,B)} \(\wedge\) \texttt{director(B,Y)}\\
        & \texttt{A(X,Y)} \(\leftarrow\) \texttt{capital(A,X)} \(\wedge\) \texttt{A(A,B)} \(\wedge\) \texttt{capital(B,Y)}\\
        & \texttt{A(X,Y)} \(\leftarrow\) \texttt{children(X,A)} \(\wedge\) \texttt{A(A,B)} \(\wedge\) \texttt{parent(B,Y)}\\
        \midrule
        \parbox[t]{1.2mm}{\multirow{3}{*}{\rotatebox[origin=c]{90}{hop=4}}} 
        
        & \texttt{A(X,Y)} \(\leftarrow\) \texttt{children(A,X)} \(\wedge\) \texttt{A(A,B)} \(\wedge\) \texttt{spouse(C,B)} \(\wedge\) \texttt{child(C,Y)}\\

         & \texttt{A(X,Y)} \(\leftarrow\) \texttt{geographic\_scope(A,X)} \(\wedge\) \texttt{A(A,B)} \(\wedge\) \texttt{headquarter(B,C)} \(\wedge\) \texttt{country(C,Y)}\\
         & \texttt{A(X,Y)} \(\leftarrow\) \texttt{country(A,X)} \(\wedge\) \texttt{citytown(B,A)} \(\wedge\) \texttt{A(B,C)} \(\wedge\) \texttt{country(C,Y)}\\

        \midrule
        \parbox[t]{1.2mm}{\multirow{3}{*}{\rotatebox[origin=c]{90}{hop=5}}} 
        & \texttt{A(X,Y)} \(\leftarrow\) \texttt{capital(A,X)} \(\wedge\) \texttt{administrative\_parent(B,A)} \(\wedge\) \texttt{A(B,C)} \(\wedge\) \texttt{state(C,D)} \(\wedge\) \texttt{capital(D,Y)}\\
        
        & \texttt{A(X,Y)} \(\leftarrow\) \texttt{currency\_used(A,X)} \(\wedge\) \texttt{country(B,A)} \(\wedge\) \texttt{A(B,C)} \(\wedge\) \texttt{country(C,D)} \(\wedge\) \texttt{currency(D,Y)}\\
        & \texttt{A(X,Y)} \(\leftarrow\) \texttt{children(X,A)} \(\wedge\) \texttt{children(B,A)} \(\wedge\) \texttt{A(B,C)} \(\wedge\) \texttt{parent(D,C)} \(\wedge\) \texttt{parent(D,Y)}\\

        \bottomrule
    \end{tabular}
    }
    \caption{\label{tab:rules}Alignment rules learned from FBDB15K.}
\end{table*}
\subsection{Discussions for Model Variants}
To investigate the effectiveness of each part of \asea, we conduct variant experiments, as detailed in Table \ref{tab:FBDBYG-Variant}.
\textbf{\textit{(\rmnum{1})}}~``\asea-Stru'' Variant: Focusing solely on graph structure, this variant demonstrates a drop in performance compared to the full \ours~model yet still surpasses other MMEA methods, indicating the practicality of the \asg~extraction algorithm and the capability of \asgnn~to effectively identify alignment rules.
\textbf{\textit{(\rmnum{2})}}~``symmetric'' Variant: Incorporating symmetric alignment rules, this variant sees further performance reductions across all metrics compared to the ``\asea-Stru'' variant, suggesting asymmetric rules also play a role in enhancing model performance.
\textbf{\textit{(\rmnum{3})}}~``W/o AMS'' Variant: By averaging multi-modal information without generating node-level attention weights, this variant shows a slight decrease in performance compared to \ours. However, it still outperforms the ``w/o value'' variant as shown in Table \ref{tab:overall-non-iter-FBDBYG}, indicating the critical importance of attribute values in EA.

This ablation study highlights the vital role of graph structure, while also showing the contributions of asymmetric rules and the Align Modal Score (AMS) module to the overall effectiveness of the \ours~model.

\subsection{Discussions for Model Depth}

\begin{figure}[!htbp]
  \centering
  \includegraphics[width=1.01\linewidth]{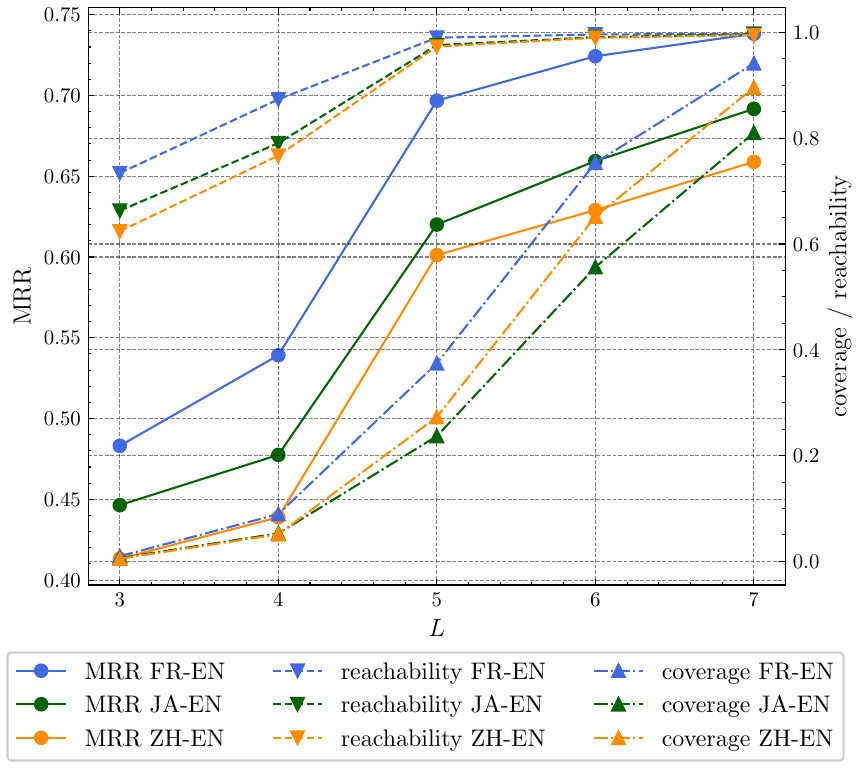}
    \caption{The MRR performance, coverage and reachability on DBP15K with different model depths ($L$).}
  \label{fig:layer}
  \vspace{-5pt}
\end{figure}
We assessed the influence of varying model depths $L$ on performance metrics MRR, coverage (i.e., the percentage of the target KG's nodes we can reach), and reachability (i.e., the proportion of target nodes that are among those reached) across different layers on the DBP15K dataset. The results, depicted in Figure \ref{fig:layer}, highlight several key insights:
\textbf{\textit{(\rmnum{1})}}~For $L<5$, performance is generally suboptimal. This is attributed to the model primarily capturing basic 1-hop alignment rules at $L=3$ and asymmetric 2:1 hop rules at $L=4$ (i.e., 2 hops in one KG and 1 hop in another), which are insufficient for comprehensive alignment.
\textbf{\textit{(\rmnum{2})}}~A notable performance boost is observed at $L=5$, indicating the effectiveness of 2-hop symmetric alignment rules.
\textbf{\textit{(\rmnum{3})}}~For $L>5$, the incorporation of more complex alignment rules leads to progressive improvements in model performance, underscoring the efficacy of the alignment rules proposed in \S~\ref{sec:rules}.

The analysis also indicates a limitation in information propagation for aligned nodes beyond $L$ hops, potentially affecting predictive accuracy. However, increasing $L$ enriches the informational content available for analysis, enhancing both the coverage and reachability of the alignment subgraph. Specifically, this is demonstrated when from $L=4$ to $L=5$, where the pronounced improvements in coverage and reachability vividly illustrate the model's enhanced performance due to the enriched information base. What's more, we find that when $L=3$, despite a mere 0.05\% coverage, it achieves a reachability of 60\%. This indicates the algorithm's ability to identify and focus on potentially aligned entities while filtering out non-relevant information.

\subsection{Visualization of interpretable}
\begin{figure}[!htbp]
  \centering
  \includegraphics[width=0.99\linewidth]{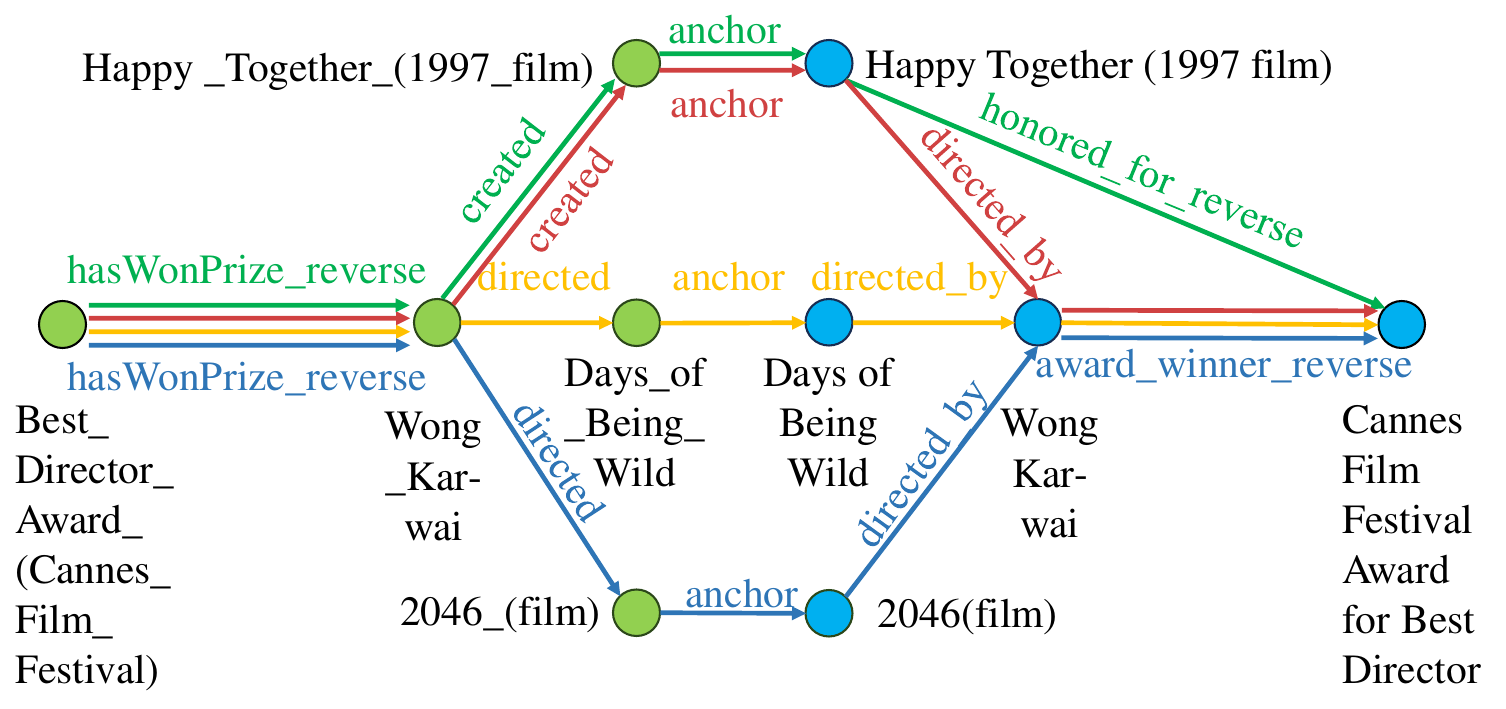}
    \caption{Visualization of learned \as}
  \label{fig:case}
  \vspace{-5pt}
\end{figure}

We can extract the alignment paths from the attention scores in \eqref{eq:alpha} in \asea for interpretation. We prune edges below a threshold to highlight identified rules. Figure \ref{fig:case} shows the alignment paths learned by \asea for a test sample. 
Given a pair of entities to align (``\textit{Best\_Director\_Award\_(Cannes\_Film\_Festival)}'', ``\textit{Cannes Film Festival Award for Best Director}''), \asea~identified four alignment paths, marked in different colors, which encompass three alignment rules. Notably, two paths (in yellow and blue) adhere to the same rule, represented as $A(e_u^1, e_v^2)\leftarrow \textit{hasWonPrize}(e_{i}^1, e_u^1) \land \textit{directed}(e_{i}^1, e_x^1) \land \textit{award\_winner}(e_v^2, e_{j}^2) \land \textit{directed\_by}(e_y^2, e_{j}^2) \land \textit{directed} = \textit{directed\_by} \land \textit{hasWonPrize} = \textit{award\_winner} \land A(e_x^1, e_y^2)$. This indicates that through the ``\textit{directed}'' and ``\textit{directed\_by}'' relations, as well as the alignment of ``\textit{Days of Being Wild}'' with ``\textit{Days\_of\_Being\_Wild}'', it can be inferred that ``\textit{Wong\_Kar-wai}'' and \textit{Wong Kar-wai}'' are the same entity. Further analysis using ``\textit{hasWonPrize}'' and ``\textit{award\_winner}'' relations concludes that the award is the same. See the Appendix for more visualization results. Additionally, Table \ref{tab:rules} enumerates some of the learned alignment rules across various hops: 
3-hop rules is the One-hop rule detailed in \S~\ref{sec:rules}. The 4-hop corresponds to an asymmetric rule where $k_1, k_2=1, 2$ (e.g., $\textit{A}(e_x,e_y)\leftarrow \textit{children}(e_a,e_x) \land \textit{A}(e_a,e_b) \land \textit{spouse}(e_c,e_b) \land \textit{child}(e_c,e_y)$), whereas the 5-hop encompasses a symmetric 2-hop rule, along with an asymmetric rule where \(k_1, k_2 = 1, 3\) (e.g., \(A(e_x,e_y) \leftarrow \textit{state\_province\_region}(e_a,e_x) \land A(e_a,e_b) \land \textit{city}(e_b,e_c) \land \textit{largestCity}(e_d,e_c) \land \textit{state}(e_d,e_y)\)). See Appendix for more learned alignment rules.

\section{Conclusion}

In conclusion, our study presents the \as~Entity Alignment (\asea) model, a significant shift from traditional embedding-based entity alignment methods in KGs. \asea, through its innovative Path-based Graph Neural Network (\asgnn) and a novel multi-modal attention mechanism, addresses interpretability challenges and enhances alignment accuracy. Our experimental results validate \asea's effectiveness in both EA and MMEA tasks, marking a notable advancement in KG entity alignment research. 
This work not only broadens the scope of entity alignment methods but also sets a foundation for future explorations in integrating multi-modal data and improving interpretability in KG applications.

\section*{Limitations}

We introduce a new paradigm for entity alignment based on logic rules. Despite our best efforts, \asea~ still
has certain limitations.

\paragraph{Limitations of Inference Complexity.}

Our method \asea~shifts away from conventional embedding-based methods, employing \as~extraction algorithms and \asgnn~for scoring to derive node-level alignment features. Despite its effectiveness in outperforming existing MMEA approaches without the need for multi-modal data, this strategy entails a higher time complexity due to the necessity of \as~extraction for each entity pair. Future work will focus on developing similarly effective models with lower time complexity.

\paragraph{Limitations in Low-Source Scenarios.}

Our approach utilizes partial anchors to connect two KGs for \as~extraction. However, in scenarios with limited anchors, it is challenging to propagate information to the aligned nodes as they are not reachable easily. Although we propose the inclusion of modal anchors as a remedy, this issue persists in cases of poor multi-modal data quality. Future efforts will explore iterative methods to improve performance in such low-source settings.

\paragraph{The Scope of Modalities.}

Our research covers three MMKG datasets, limited to text and image modalities. However, MMKGs also include other modalities like temporal information. Although our Align modal score is designed to handle such diversity theoretically, we lack empirical validation. Moving forward, we plan to investigate more modalities to assess our model's limits and increase its adaptability across various data types.

\section*{Ethical Considerations}
Our experiments and model training strictly use publicly available datasets (as detailed in
Appendix), mitigating ethical issues regarding privacy, confidentiality, or the misuse of personal biological information.


\bibliography{anthology,custom}

\appendix

\clearpage

\section{Appendix}
\label{sec:appendix}
\subsection{Dataset Statistics}
The detailed statistics of our dataset are delineated in Table \ref{tab:dataset}, aligning with \citet{DBLP:conf/mm/ChenCZGFHZGPSC23,DBLP:conf/semweb/ChenGFZCPLCZ23,DBLP:journals/corr/abs-2305-14651}. It is important to note that our dataset includes a collection of pre-aligned entity pairs. These pairs are systematically divided into a training set (denoted as seed alignments $\mathcal{S}$) and a testing set (denoted as $\mathcal{S}_{te}$), under the specified seed alignment ratio ($R_{sa}$).
\begin{table*}[!htbp]
    \centering
    \vspace{-3pt}
    \renewcommand\arraystretch{1.0}
    \resizebox{0.77\linewidth}{!}{
    \begin{tabular}{@{}l|c|cccccccc@{}}
        \toprule
        \makebox[2.5cm][c]{Dataset} & KG & \# Ent. & \# Rel. & \# Attr. & \# Rel. Triples & \# Attr. Triples & \# Image & \# EA pairs \\
        \midrule
        \multirow{2}*{FBDB15K} & FB15K & 14,951 & 1,345 & 116 & 592,213 & 29,395 & 13,444 & \multirow{2}*{12,846} \\
        & DB15K & 12,842 & 279 & 225 & 89,197 & 48,080 & 12,837 \\
        \midrule
        \multirow{2}*{FBYG15K} & FB15K & 14,951 & 1,345 & 116 & 592,213 & 29,395 & 13,444 & \multirow{2}*{11,199} \\
        & YAGO15K & 15,404 & 32 & 7 & 122,886 & 23,532 & 11,194 \\
        \midrule
        \multirow{2}*{DBP15K$_{\text{ZH-EN}}$} & ZH {\footnotesize (Chinese)} & 19,388 & 1,701 & 8,111 & 70,414 & 248,035 & 15,912 & \multirow{2}*{15,000} \\
        & EN {\footnotesize (English)} & 19,572 & 1,323 & 7,173 & 95,142 & 343,218 & 14,125 \\
        \midrule
        \multirow{2}*{DBP15K$_{\text{JA-EN}}$} & JA {\footnotesize (Japanese)} & 19,814 & 1,299 & 5,882 & 77,214 & 248,991 & 12,739 & \multirow{2}*{15,000} \\
        & EN {\footnotesize (English)} & 19,780 & 1,153 & 6,066 & 93,484 & 320,616 & 13,741 \\
        \midrule
        \multirow{2}*{DBP15K$_{\text{FR-EN}}$} & FR {\footnotesize (French)} & 19,661 & 903 & 4,547 & 105,998 & 273,825 & 14,174 & \multirow{2}*{15,000} \\
        & EN {\footnotesize (English)} & 19,993 & 1,208 & 6,422 & 115,722 & 351,094 & 13,858 \\
        
        \midrule
        \multirow{2}*{OpenEA$_{\text{EN-FR}}$} & EN {\footnotesize (English)} & 15,000 & 267 & 308 & 47,334 & 73,121 & 15,000 & \multirow{2}*{15,000} \\
        & FR {\footnotesize (French)} & 15,000 & 210 & 404 & 40,864 & 67,167 & 15,000 \\
        \midrule
        \multirow{2}*{OpenEA$_{\text{EN-DE}}$} & EN {\footnotesize (English)} & 15,000 & 215 & 286 & 47,676 & 83,755 & 15,000 & \multirow{2}*{15,000} \\
        & DE (German) & 15,000 & 131 & 194 & 50,419 & 156,150 & 15,000 \\
                \midrule
        \multirow{2}*{OpenEA$_{\text{D-W-V1}}$} & DBpedia & 15,000 & 248 & 342 & 38,265 & 68,258 & 15,000 & \multirow{2}*{15,000} \\
        & Wikidata & 15,000 & 169 & 649 & 42,746 & 138,246 & 15,000 \\
        \midrule
        \multirow{2}*{OpenEA$_{\text{D-W-V2}}$} & DBpedia & 15,000 & 167 & 175 & 73,983 & 66,813 & 15,000 & \multirow{2}*{15,000} \\
        & Wikidata & 15,000 & 121 & 457 & 83,365 & 175,686 & 15,000 \\
        \bottomrule
    \end{tabular}
    }
    \caption{\label{tab:dataset}Statistics for three datasets, where ``EA pairs'' refers to the pre-aligned entity pairs.}
    \vspace{-3pt}
    \end{table*}
\subsection{Baseline Details}
\subsubsection{Comparision Methods}

For traditional EA methods, we select IPTransE \cite{DBLP:conf/ijcai/ZhuXLS17}, GCN-Align \cite{DBLP:conf/emnlp/WangLLZ18}, KECG \cite{DBLP:conf/emnlp/LiCHSLC19}, AlignEA \cite{DBLP:conf/ijcai/SunHZQ18}, SEA \cite{DBLP:conf/www/PeiYHZ19} and IMUSE \cite{DBLP:conf/dasfaa/HeLQ0LZ0ZC19},MUGNN \cite{DBLP:conf/acl/CaoLLLLC19},AliNet \cite{DBLP:conf/aaai/SunW0CDZQ20} as baselines. For Multi-modal methods, we select PoE \cite{DBLP:conf/esws/LiuLGNOR19}, MMEA \cite{DBLP:conf/ksem/ChenLWXWC20}, EVA \cite{DBLP:conf/aaai/0001CRC21}, MCLEA \cite{DBLP:conf/coling/LinZWSW022}, MSNEA \cite{DBLP:conf/kdd/ChenL00WYC22}, MEAformer \cite{DBLP:conf/mm/ChenCZGFHZGPSC23}, ACK-MMEA \cite{DBLP:conf/www/LiGLJWSL23}, UMAEA \cite{DBLP:conf/semweb/ChenGFZCPLCZ23}. For a clear comparison, all the baselines are not iterable and do not use the surface information. 
\subsubsection{Traditional methods}
Traditional methods primarily utilize relational modalities, with some incorporating attribute modalities. Key methods in this category include:

IPTransE \cite{DBLP:conf/ijcai/ZhuXLS17}: It iteratively expands a set of softly aligned entity pairs, using a shared parameter strategy for different KGs.

GCN-Align \cite{DBLP:conf/emnlp/WangLLZ18}: This approach integrates structural and attribute information through graph convolutional networks.

BootEA \cite{DBLP:conf/ijcai/SunHZQ18}: It labels probable alignments as training data, reducing error accumulation through an iterative alignment method.
 
SEA \cite{DBLP:conf/www/PeiYHZ19}: This method leverages both labeled and abundant unlabeled entity information for alignment, incorporating adversarial training to account for differences in point degrees.

IMUSE \cite{DBLP:conf/dasfaa/HeLQ0LZ0ZC19}: It employs a bivariate regression model for combining alignment results, focusing on relation and attribute similarities.

\subsubsection{Multi-modal methods}
Multi-modal methods involve relational, attribute, and visual modalities. Prominent methods include:

EVA \cite{DBLP:conf/aaai/0001CRC21}: Utilizes visual semantic representations for unsupervised entity alignment in heterogeneous KGs, excelling in aligning long-tail entities by creating comprehensive entity representations.

MMEA \cite{DBLP:conf/ksem/ChenLWXWC20}: Generates entity representations from relational, visual, and numerical knowledge, and then integrates these modalities.

MCLEA \cite{DBLP:conf/coling/LinZWSW022}: Proposes a multi-modal contrastive learning framework for entity representation learning across different modalities.

MSNEA \cite{DBLP:conf/kdd/ChenL00WYC22}: Introduces a multi-modal siamese network approach for entity representation learning.

MEAformer \cite{DBLP:conf/mm/ChenCZGFHZGPSC23}: Employs a multi-modal transformer-based approach for learning entity representations.

ACK-MMEA \cite{DBLP:conf/www/LiGLJWSL23}: Focuses on contextual gap problems, and introduces an attribute-consistent KG entity alignment framework considering multiple modalities.

UMAEA \cite{DBLP:conf/semweb/ChenGFZCPLCZ23}: A robust multi-modal entity alignment method designed to handle uncertainties and ambiguities in visual modalities.
\subsection{Implementation Details}
We tune the learning rate in [$10^{-4}$,$10^{-2}$], weight decay in [$10^{-5}$,$10^{-2}$], dropout rate in [$0$,$0.3$], batch size in \{4, 8, 16\}, dimension $d$ in \{32,48,64\}, Adam \cite{DBLP:journals/corr/KingmaB14} is used as the optimizer, the training epoch is 50. For the GNN layer $L$, we use $L=5$ for MMKG and $L=7$ for DBP15K and Multi-OpenEA depending on the density of KG. All experiments were conducted on Tesla V100. We employ the Gated Recurrent Unit (GRU)~\cite{DBLP:journals/corr/ChungGCB14} as our RNN model. Our work is based on the open-source codebase of RED-GNN\footnote{\url{https://github.com/LARS-research/RED-GNN}.} and MEAformer\footnote{\url{https://github.com/zjukg/MEAformer}. MIT license.}. For other baselines AnyBURL\footnote{\url{https://web.informatik.uni-mannheim.de/AnyBURL}. 3-clause BSD license.} and Neural LP\footnote{\url{https://github.com/kexinyi71/Neural-LP}. MIT license.}, we use their open-source codebase.
\subsection{Complete experiment results}
We provide the complete results of all experiments on MMKG, DBP15K, and Multi-OpenEA datasets as shown in Tables \ref{tab:overall-non-iter-FBDBYG}, \ref{tab:dbp15k-all}, and \ref{tab:moea-all}.
\begin{table*}[ht]
    \centering
    \renewcommand\arraystretch{1.0}
    
    \resizebox{\textwidth}{!}{
\begin{tabular}{@{}l|l|ccc|ccc|ccc|ccc|ccc|ccc}
        \toprule
        & \multirow{2}*{\makebox[1.5cm][c]{Models}} & \multicolumn{3}{c|}{FBDB15K(20\%)} & \multicolumn{3}{c|}{FBDB15K(50\%)} & \multicolumn{3}{c|}{FBDB15K(80\%)} & \multicolumn{3}{c|}{FBYG15K(20\%)} & \multicolumn{3}{c|}{FBYG15K(50\%)} & \multicolumn{3}{c}{FBYG15K(80\%)} \\
        & & {\scriptsize H@1} & {\scriptsize H@10} & {\scriptsize MRR} & {\scriptsize H@1} & {\scriptsize H@10} & {\scriptsize MRR} & {\scriptsize H@1} & {\scriptsize H@10} & {\scriptsize MRR} & {\scriptsize H@1} & {\scriptsize H@10} & {\scriptsize MRR} & {\scriptsize H@1} & {\scriptsize H@10} & {\scriptsize MRR} & {\scriptsize H@1} & {\scriptsize H@10} & {\scriptsize MRR} \\
        \midrule
        \parbox[t]{2mm}{\multirow{8}{*}{\rotatebox[origin=c]{90}{EA}}} 
        & IPTransE
        & .065 & .215 & .094 & .210 & .421 & .283 & .403 & .627 & .469 & .047 & .169 & .084 & .201 & .369 & .248 & .401 & .602 & .458 \\
        & GCN-align  
        & .053 & .174 & .087 & .226 & .435 & .293 & .414 & .635 & .472 & .081 & .235 & .153 & .233 & .424 & .294 & .406 & .643 & .477 \\
        & SEA  
        & .170 & .425 & .255 & .373 & .657 & .470 & .512 & .784 & .505 & .141 & .371 & .218 & .294 & .577 & .388 & .514 & .773 & .605 \\
        & IMUSE  
        & .176 & .435 & .264 & .309 & .576 & .400 & .457 & .726 & .551 & .081 & .257 & .142 & .398 & .601 & .469 & .512 & .707 & .581 \\
        & \CC Neural LP$\ast$  
        & \CC.098 & \CC.297 & \CC.161 & \CC.233 & \CC.552 & \CC.339 & \CC.314 & \CC.604 & \CC.412 & \CC.106 & \CC.285 & \CC.163 & \CC.281 & \CC.531 & \CC.362 & \CC.343 & \CC.631 & \CC.441 \\
        & \CC AnyBURL$\ast$  
        & \CC.213 & \CC.515 & \CC.312 & \CC.386 & \CC.698 & \CC.491 & \CC.437 & \CC.727 & \CC.533 & \CC.210 & \CC.519 & \CC.311 & \CC.356 & \CC.662 & \CC.459 & \CC.397 & \CC.678 & \CC.489 \\
        
        & \CC RED-GNN$\ast$  
        & \CC.482 & \CC.687 & \CC.552 & \CC.692 & \CC.843 & \CC.746 & \CC.765 & \CC.882 & \CC.817 & \CC.464 & \CC.659 & \CC.532 & \CC.647 & \CC.815 & \CC.707 & \CC.728 & \CC.871 & \CC.779 \\
        & \CC\textbf{\asea-Stru}  
        & \CC\textbf{.506} & \CC\textbf{.706} & \CC\textbf{.575} & \CC\textbf{.711} & \CC\textbf{.860} & \CC\textbf{.765} & \CC\textbf{.791} & \CC\textbf{.897} & \CC\textbf{.828} & \CC\textbf{.479} & \CC\textbf{.681} & \CC\textbf{.548} & \CC\textbf{.659} & \CC\textbf{.827} & \CC\textbf{.724} & \CC\textbf{.743} & \CC\textbf{.888} & \CC\textbf{.796} \\
        \midrule
        \parbox[t]{2mm}{\multirow{8}{*}{\rotatebox[origin=c]{90}{MMEA}}}
        & MMEA  
        & .265 & .541 & .357 & .417 & .703 & .512 & .590 & .869 & .685 & .234 & .480 & .317 & .403 & .645 & .486 & .598 & .839 & .682 \\
        & EVA  
        & .199 & .448 & .283 & .334 & .589 & .422 & .484 & .696 & .563 & .153 & .361 & .224 & .311 & .534 & .388 & .491 & .692 & .565 \\
        & MSNEA  
        & .114 & .296 & .175 & .288 & .590 & .388 & .518 & .779 & .613 & .103 & .249 & .153 & .320 & .589 & .413 & .531 & .778 & .620 \\
        & MCLEA  
        & .295 & .582 & .393 & .555 & .784 & .637 & .735 & .890 & .790 & .254 & .484 & .332 & .501 & .705 & .574 & .667 & .824 & .722 \\
        & ACK-MMEA  
        & .304 & .549 & .387 & .560 & .736 & .624 & .682 & .874 & .752 & .289 & .496 & .360 & .535 & .699 & .593 & .676 & .864 & .744 \\
        & MEAformer  
        & .434 & .728 & .534 & .625 & .847 & .704 & .773 & .918 & .825 & .325 & .598 & .416 & .560 & .780 & .640 & .705 & .874 & .768 \\
        & \CC\textbf{\ours~{\footnotesize{w/o value}}}  
        & \CC\textbf{.516} & \CC\textbf{.718} & \CC\textbf{.586} & \CC\textbf{.725} & \CC\textbf{.867} & \CC\textbf{.777} & \CC\textbf{.800} & \CC\textbf{.908} & \CC\textbf{.839} & \CC\textbf{.493} & \CC\textbf{.696} & \CC\textbf{.563} & \CC\textbf{.671} & \CC\textbf{.835} & \CC\textbf{.730} & \CC\textbf{.756} & \CC\textbf{.893} & \CC\textbf{.807} \\
        & \CC\textbf{\ours}  
        & \CC\textbf{.628} & \CC\textbf{.799} & \CC\textbf{.689} & \CC\textbf{.794} & \CC\textbf{.899} & \CC\textbf{.833} & \CC\textbf{.852} & \CC\textbf{.927} & \CC\textbf{.881} & \CC\textbf{.717} & \CC\textbf{.848} & \CC\textbf{.776} & \CC\textbf{.842} & \CC\textbf{.920} & \CC\textbf{.879} & \CC\textbf{.902} & \CC\textbf{.965} & \CC\textbf{.926} \\
        \bottomrule
    \end{tabular}
    }
    \caption{\cz{Performance comparison on FBDB15K and FBYG15K datasets at various reference $R_{sa}$ for training, between EA and MMEA models. ``\asea-MM'' indicates our complete model that uses multi-modal information(abbreviated as MM), while ``\asea-Stru'' indicates reliance on graph structure alone (i.e., triples), ``w/o value'' denotes the exclusion of attribute value.} ``$\ast$'' indicates the reasoning method adapted in our framework to EA task.}
    \label{tab:overall-non-iter-FBDBYG}
\end{table*}
\begin{table*}[!htbp]
    \centering
    \tabcolsep=0.3cm
    \renewcommand\arraystretch{1.0}
    
    \resizebox{1\linewidth}{!}{
    \begin{tabular}{@{}l|l|ccc|ccc|ccc}
        \toprule
        & \multirow{2}*{\makebox[2cm][c]{Models}} & \multicolumn{3}{c|}{DBP15K$_{\text{ZH-EN}}$} & \multicolumn{3}{c|}{DBP15K$_{\text{JA-EN}}$} & \multicolumn{3}{c}{DBP15K$_{\text{FR-EN}}$} \\
        & & {\scriptsize H@1} & {\scriptsize H@10} & {\scriptsize MRR} & {\scriptsize H@1} & {\scriptsize H@10} & {\scriptsize MRR} & {\scriptsize H@1} & {\scriptsize H@10} & {\scriptsize MRR} \\
        \midrule
        \parbox[t]{2mm}{\multirow{5}{*}{\rotatebox[origin=c]{90}{EA}}}
        &AlignEA {\footnotesize \cite{DBLP:conf/ijcai/SunHZQ18}} & 
        .472 & .792 & .581 & .448 & .789 & .563 & .481 & .824 & .599 \\
        &KECG {\footnotesize {\cite{DBLP:conf/emnlp/LiCHSLC19}}} &
        .478 & .835 & .598 & .490 & .844 & .610 & .486 & .851 & .610 \\
        &MUGNN {\footnotesize \cite{DBLP:conf/acl/CaoLLLLC19}} &
        .494 & .844 & .611 &  .501 & .857 & .621 & .495 & .870 & .621 \\
        &AliNet {\footnotesize {\cite{DBLP:conf/aaai/SunW0CDZQ20}}} &
        \underline{.539} & \underline{.826} & \underline{.628} & \underline{.549} & \underline{.831} & \underline{.645} & \underline{.552} & \underline{.852} & \underline{.657} \\
        &\CC\textbf{\asea-Stru}  &
        \CC\textbf{.560} & \CC\textbf{.844} & \CC\textbf{.660} & \CC\textbf{.595} & \CC\textbf{.866} & \CC\textbf{.690} & \CC\textbf{.653} & \CC\textbf{.906} & \CC\textbf{.745} \\
        \midrule
        \parbox[t]{2mm}{\multirow{6}{*}{\rotatebox[origin=c]{90}{MMEA}}}
        &MSNEA {\footnotesize {\cite{DBLP:conf/kdd/ChenL00WYC22}}} & .609 & .831 & .685 & .541 & .776 & .620 & .557 & .820 & .643 \\
        &EVA {\footnotesize \cite{DBLP:conf/aaai/0001CRC21}} &
        {.683} & {.906} & {.762} & {.669} & {.904} & {.752} & {.686} & .928 & {.771} \\
        &MCLEA {\footnotesize {\cite{DBLP:conf/coling/LinZWSW022}}}  &
        .726 & .922 & .796 & .719 & .915 & .789 & .719 & .918 & .792 \\
        &MEAformer {\footnotesize {\cite{DBLP:conf/mm/ChenCZGFHZGPSC23}}}&
        .771 & .951 & .835 & .764 & .959 & .834 & .770 & .961 & .841 \\
        &UMAEA {\footnotesize {\cite{DBLP:conf/semweb/ChenGFZCPLCZ23}}}  & \underline{.800} & \underline{.962} & \underline{.860} & \underline{.801} & \underline{.967} & \underline{.862} & \underline{.818} & \underline{.973} & \underline{.877} \\
        &\CC\textbf{\ours}  &
        \CC\textbf{.815} & \CC\textbf{.941} & \CC\textbf{.862} & \CC\textbf{.849} & \CC\textbf{.956} & \CC\textbf{.889} & \CC\textbf{.911} & \CC\textbf{.982} & \CC\textbf{.938} \\
        
        \bottomrule
    \end{tabular}
    }
    \caption{\cz{Complete results on DPB15K \cite{DBLP:conf/semweb/SunHL17} datasets.} 
    }
    \label{tab:dbp15k-all}
\end{table*}
\begin{table*}[!htbp]
    \centering
    \tabcolsep=0.3cm
    \renewcommand\arraystretch{1.0}
    
    \resizebox{0.99\linewidth}{!}{
    \begin{tabular}{@{}l|ccc|ccc|ccc|ccc}
        \toprule
        \multirow{2}*{\makebox[2cm][c]{Models}} & \multicolumn{3}{c|}{OpenEA$_{\text{EN-FR}}$} & \multicolumn{3}{c|}{OpenEA$_{\text{EN-DE}}$} & \multicolumn{3}{c|}{OpenEA$_{\text{D-W-V1}}$} & \multicolumn{3}{c}{OpenEA$_{\text{D-W-V2}}$} \\
        & {\scriptsize H@1} & {\scriptsize H@10} & {\scriptsize MRR} & {\scriptsize H@1} & {\scriptsize H@10} & {\scriptsize MRR} & {\scriptsize H@1} & {\scriptsize H@10} & {\scriptsize MRR} & {\scriptsize H@1} & {\scriptsize H@10} & {\scriptsize MRR} \\
        \midrule
        MSNEA* {\footnotesize {\cite{DBLP:conf/kdd/ChenL00WYC22}}} 
        & .692 & .813 & .734 & .753 & .895 & .804 & .800 & .874 & .826 & .838 & .940 & .873 \\
        EVA* {\footnotesize \cite{DBLP:conf/aaai/0001CRC21}} 
        & {.785} & {.932} & {.836} & {.922} & {.983} & {.945} & {.858} & {.946} & {.891} & .890 & .981 & .922 \\
        MCLEA* {\footnotesize {\cite{DBLP:conf/coling/LinZWSW022}}}  
        & .819 & .943 & .864 & .939 & .988 & .957 & .881 & .955 & .908 & .928 & .983 & .949 \\
        UMAEA* {\footnotesize {\cite{DBLP:conf/semweb/ChenGFZCPLCZ23}}}  
        & \underline{.848} & \underline{.966} & \underline{.891} & \underline{.956} & \underline{.994} & \underline{.971} & \underline{\textbf{.904}} & \underline{\textbf{.971}} & \underline{\textbf{.930}} & \underline{.948} & \underline{\textbf{.996}} & \underline{.967} \\
        \CC\textbf{\ours}  
        & \CC\textbf{.911} & \CC\textbf{.968} & \CC\textbf{.932} & \CC\textbf{.980} & \CC\textbf{.996} & \CC\textbf{.986} & \CC{.898} & \CC{.968} & \CC{.924} & \CC\textbf{.958} & \CC{.994} & \CC\textbf{.972} \\

        \bottomrule
    \end{tabular}
    }
    \caption{Complete results on four Multi-OpenEA datasets.  
    }
    \label{tab:moea-all}
\end{table*}
\subsection{\as~Extracting Algorithm}

We have developed various versions of the extracting algorithm, including a non-merging version, a merging version, and a version focused solely on symmetric rules.

The non-merging version, as delineated in Algorithm \ref{ag:1}, extracts the \asg~between \(e_u\) and \(e_v\).

The merging version, outlined in Algorithm \ref{ag:2}, extracts a combined alignment subgraph from entity \(e_u\) to all entities in $\mathcal{G}_2$.

The version concentrating only on symmetric rules filters the rules accordingly, ensuring that the \as~comprises only paths based on symmetric rules, as shown in Algorithm \ref{ag:3}.

\begin{algorithm*}
\caption{\asg~Extraction}
\begin{algorithmic}[1]
\label{ag:1}
\REQUIRE $\mathcal{G}_1 = \{\mathcal{E}_1, \mathcal{R}_1, \mathcal{T}_1\}$, $\mathcal{G}_2 = \{\mathcal{E}_2, \mathcal{R}_2, \mathcal{T}_2\}$, Anchor Links $A$, Node pair $(e_u,e_v)$, hop $K$
\ENSURE $K$-hop \asg~$g^K(e_u,e_v)$ for node pair $(e_u,e_v)$

\STATE Initialize $\mathcal{E}_s^0 \leftarrow \{e_u\}$ , $\mathcal{E}_t^0 \leftarrow \{e_v\}$, $\mathcal{T} \leftarrow \mathcal{T}_1 \cup \mathcal{T}_2 \cup A \cup \mathcal{T}_1^{rev} \cup \mathcal{T}_2^{rev} \cup A^{rev}$

\FOR{$i = 1$ to $K$}
    \STATE $\mathcal{E}_s^i \leftarrow \{e \mid (e_u, r, e) \in \mathcal{T} , e_u \in \mathcal{E}_s^{i-1}\}$
    \STATE $\mathcal{T}^i \leftarrow \{(e_u, r, e) \mid (e_u, r, e) \in \mathcal{T} , e_u \in \mathcal{E}_s^{i-1}\}$
\ENDFOR
\FOR{$i = 1$ to $K$}
    \STATE $\mathcal{E}_t^i \leftarrow \{e \mid (e, r, e_v) \in \mathcal{T}^{K-i} , e_v \in \mathcal{E}_t^{i-1}\} \cup \mathcal{E}_t^{i-1}$
    \STATE $\mathcal{T}_t^i \leftarrow \{(e, r, e_v) \mid (e, r, e_v) \in \mathcal{T}^{K-i} , e_v \in \mathcal{E}_t^{i-1}\}$
\ENDFOR
\RETURN $\mathcal{T}_t^0 \cup \cdots \cup \mathcal{T}_t^K$
\end{algorithmic}
\end{algorithm*}

\begin{algorithm*}
\caption{Merged \asg~Extraction}
\begin{algorithmic}[1]
\label{ag:2}
\REQUIRE $\mathcal{G}_1 = \{\mathcal{E}_1, \mathcal{R}_1, \mathcal{T}_1\}$, $\mathcal{G}_2 = \{\mathcal{E}_2, \mathcal{R}_2, \mathcal{T}_2\}$, Anchor Links $A$, Node $e_u$, hop $K$
\ENSURE $K$-hop \asg~$g_u^K$ for Node $e_u$

\STATE Initialize $E_s^0 \leftarrow \{e_u\}$, $\textbf{E}_t^0 \leftarrow\left\{\begin{array}{ll}
  \mathcal{E}_1, &  e_u \in \mathcal{E}_2 \\
  \mathcal{E}_2, &  e_u \in \mathcal{E}_1
  \end{array}\right.$ ,$\mathcal{T} \leftarrow \mathcal{T}_1 \cup \mathcal{T}_2 \cup A \cup \mathcal{T}_1^{rev} \cup \mathcal{T}_2^{rev} \cup A^{rev}$

\FOR{$i = 1$ to $K$}
    \STATE $\mathcal{E}_s^i \leftarrow \{e \mid (e_u, r, e) \in \mathcal{T} , e_u \in \mathcal{E}_s^{i-1}\}$
    \STATE $\mathcal{T}^i \leftarrow \{(e_u, r, e) \mid (e_u, r, e) \in \mathcal{T} , e_u \in \mathcal{E}_s^{i-1}\}$
\ENDFOR
\FOR{$i = 1$ to $K$}
    \STATE $\mathcal{E}_t^i \leftarrow \{e \mid (e, r, e_u) \in \mathcal{T}^{K-i} , e_u \in \mathcal{E}_t^{i-1}\} \cup \mathcal{E}_t^{i-1}$
    \STATE $T_t^i \leftarrow \{(e, r, e_u) \mid (e, r, e_u) \in \mathcal{T}^{K-i} , e_u \in \mathcal{E}_t^{i-1}\}$
\ENDFOR
\RETURN $\mathcal{T}_t^0 \cup \cdots \cup \mathcal{T}_t^K$
\end{algorithmic}
\end{algorithm*}

\begin{algorithm*}
\caption{Symmetric \asg~Extraction}
\begin{algorithmic}[1]
\label{ag:3}
\REQUIRE $\mathcal{G}_1 = \{\mathcal{E}_1, \mathcal{R}_1, \mathcal{T}_1\}$, $\mathcal{G}_2 = \{\mathcal{E}_2, \mathcal{R}_2, \mathcal{T}_2\}$, Anchor Links $A$, Node $e_u$, hop $K$
\ENSURE $K$-hop \as~$g_u^K$ for Node $e_u$

\STATE Initialize $\mathcal{E}_s^0 \leftarrow \{e_u\}$, $\mathcal{E}_t^0 \leftarrow\left\{\begin{array}{ll}
  \mathcal{E}_1, &  e_u \in \mathcal{E}_2 \\
  \mathcal{E}_2, &  e_u \in \mathcal{E}_1
  \end{array}\right.$, $\mathcal{T} \leftarrow \mathcal{T}_1 \cup \mathcal{T}_2 \cup A \cup \mathcal{T}_1^{rev} \cup \mathcal{T}_2^{rev} \cup A^{rev}$

\FOR{$i = 1$ to $K$}
    \STATE $\mathcal{E}_s^i \leftarrow \{e \mid (e_u, r, e) \in \mathcal{T} , e_u \in \mathcal{E}_s^{i-1}\}$
    \STATE $\mathcal{T}^i \leftarrow \{(e_u, r, e) \mid (e_u, r, e) \in \mathcal{T} , e_u \in \mathcal{E}_s^{i-1}\}$
\ENDFOR
\FOR{$i = 1$ to $K$}
    \STATE $\mathcal{E}_t^i \leftarrow \{e \mid (e, r, e_u) \in \mathcal{T}^{K-i} , e_u \in \mathcal{E}_t^{i-1}\} \cup \mathcal{E}_t^{i-1}$
    
    \STATE $\mathcal{T}_t^i \leftarrow \{(e, r, e_u) \mid (e, r, e_u) \in \mathcal{T}^{K-i} , e_u \in \mathcal{E}_t^{i-1}\}$
    \IF{$i < K//2$}
        \STATE $\mathcal{E}_t^i \leftarrow \mathcal{E}_t^i\cap \mathcal{E}_2$
    \ELSE
        \STATE $\mathcal{E}_t^i \leftarrow \mathcal{E}_t^i \cap \mathcal{E}_1$
    \ENDIF
\ENDFOR
\RETURN $\mathcal{T}_t^0 \cup \cdots \cup \mathcal{T}_t^K$
\end{algorithmic}
\end{algorithm*}
\subsection{More Visualization of Learned \asg and alignment rules}
\begin{figure*}[!htbp]
  \centering
  \includegraphics[width=0.9\linewidth]{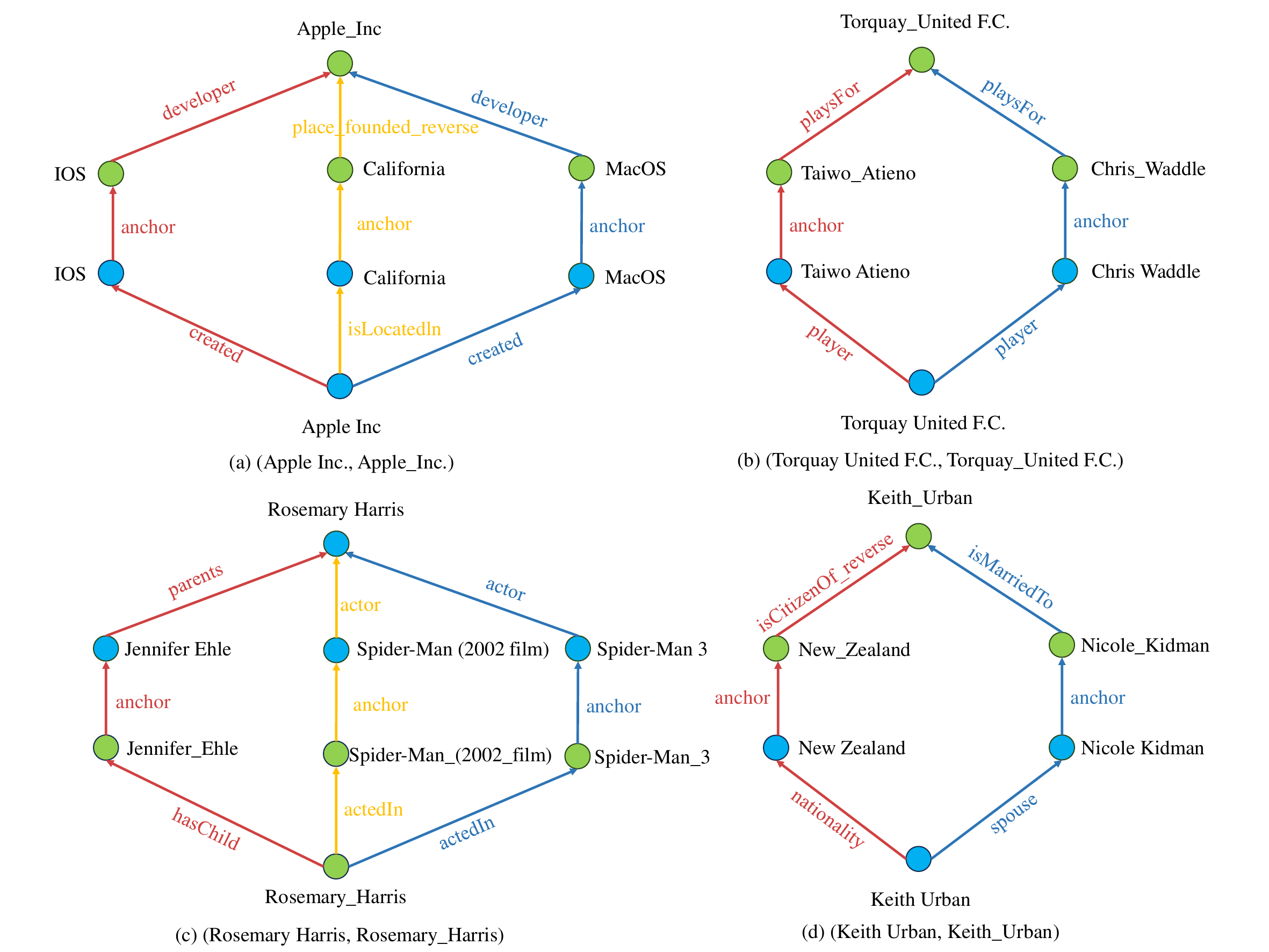}
    \caption{More visualization of learned \asg}
  \label{fig:add-vis}
  \vspace{-5pt}
\end{figure*}

Figure \ref{fig:add-vis} visualizes more learned \asg~with alignment paths on different samples. We also provide more learned rules in Table \ref{tab:rules-all}.
\begin{table*}[ht]
    \centering
    \renewcommand\arraystretch{1.0}
    \resizebox{0.9\linewidth}{!}{
    \begin{tabular}{@{}l|l|ccc|ccc}
        \midrule
        \parbox[t]{1.2mm}{\multirow{5}{*}{\rotatebox[origin=c]{90}{hop=3}}} 
        & \texttt{A(X,Y)} \(\leftarrow\) \texttt{directed\_by(A,X)} \(\wedge\) \texttt{A(A,B)} \(\wedge\) \texttt{director(B,Y)}\\
        & \texttt{A(X,Y)} \(\leftarrow\) \texttt{capital(A,X)} \(\wedge\) \texttt{A(A,B)} \(\wedge\) \texttt{capital(B,Y)}\\
        & \texttt{A(X,Y)} \(\leftarrow\) \texttt{children(X,A)} \(\wedge\) \texttt{A(A,B)} \(\wedge\) \texttt{parent(B,Y)}\\
        & \texttt{A(X,Y)} \(\leftarrow\) \texttt{written\_by(A,X)} \(\wedge\) \texttt{A(A,B)} \(\wedge\) \texttt{writer(B,Y)}\\
        & \texttt{A(X,Y)} \(\leftarrow\) \texttt{player(A,X)} \(\wedge\) \texttt{A(A,B)} \(\wedge\) \texttt{draftTeam(Y,B)}\\
        \midrule
        \parbox[t]{1.2mm}{\multirow{5}{*}{\rotatebox[origin=c]{90}{hop=4}}} 
        
        & \texttt{A(X,Y)} \(\leftarrow\) \texttt{children(A,X)} \(\wedge\) \texttt{A(A,B)} \(\wedge\) \texttt{spouse(C,B)} \(\wedge\) \texttt{child(C,Y)}\\

        & \texttt{A(X,Y)} \(\leftarrow\) \texttt{capital(X,A)} \(\wedge\) \texttt{county\_seat(B,A)} \(\wedge\) \texttt{A(B,C)} \(\wedge\) \texttt{state(C,Y)}\\

        & \texttt{A(X,Y)} \(\leftarrow\) \texttt{location(A,X)} \(\wedge\) \texttt{school(A,B)} \(\wedge\) \texttt{A(B,C)} \(\wedge\) \texttt{city(C,Y)}\\
         & \texttt{A(X,Y)} \(\leftarrow\) \texttt{geographic\_scope(A,X)} \(\wedge\) \texttt{A(A,B)} \(\wedge\) \texttt{headquarter(B,C)} \(\wedge\) \texttt{country(C,Y)}\\
         & \texttt{A(X,Y)} \(\leftarrow\) \texttt{country(A,X)} \(\wedge\) \texttt{citytown(B,A)} \(\wedge\) \texttt{A(B,C)} \(\wedge\) \texttt{country(C,Y)}\\

        \midrule
        \parbox[t]{1.2mm}{\multirow{5}{*}{\rotatebox[origin=c]{90}{hop=5}}} 
        & \texttt{A(X,Y)} \(\leftarrow\) \texttt{capital(A,X)} \(\wedge\) \texttt{administrative\_parent(B,A)} \(\wedge\) \texttt{A(B,C)} \(\wedge\) \texttt{state(C,D)} \(\wedge\) \texttt{capital(D,Y)}\\
        
        & \texttt{A(X,Y)} \(\leftarrow\) \texttt{parents(A,X)} \(\wedge\) \texttt{award\_nominee(B,A)} \(\wedge\) \texttt{A(B,C)} \(\wedge\) \texttt{starring(C,D)} \(\wedge\) \texttt{parent(D,Y)}\\
        & \texttt{A(X,Y)} \(\leftarrow\) \texttt{state\_province\_region(A,X)} \(\wedge\) \texttt{A(A,B)} \(\wedge\) \texttt{city(B,C)} \(\wedge\) \texttt{largestCity(D,C)} \(\wedge\) \texttt{state(D,Y)}\\
        & \texttt{A(X,Y)} \(\leftarrow\) \texttt{currency\_used(A,X)} \(\wedge\) \texttt{country(B,A)} \(\wedge\) \texttt{A(B,C)} \(\wedge\) \texttt{country(C,D)} \(\wedge\) \texttt{currency(D,Y)}\\
        & \texttt{A(X,Y)} \(\leftarrow\) \texttt{children(X,A)} \(\wedge\) \texttt{children(B,A)} \(\wedge\) \texttt{A(B,C)} \(\wedge\) \texttt{parent(D,C)} \(\wedge\) \texttt{parent(D,Y)}\\

        \bottomrule
    \end{tabular}
    }
    \caption{\label{tab:rules-all}More alignment rules learned from FBDB15K.}
\end{table*}

\end{document}